\pdfoutput=1

\documentclass[11pt]{article}

\usepackage[final]{acl}

\usepackage{times}
\usepackage{latexsym}
\usepackage{amsfonts}
\usepackage{amsmath}
\usepackage[table]{xcolor}

\usepackage[T1]{fontenc}

\usepackage[utf8]{inputenc}

\usepackage{microtype}

\usepackage{inconsolata}

\usepackage{graphicx}

\usepackage{booktabs}       

\usepackage{multirow}
\usepackage[capitalise]{cleveref}
\crefformat{section}{\S#2#1#3} 
\crefformat{subsection}{\S#2#1#3}
\crefformat{subsubsection}{\S#2#1#3}

\usepackage{todonotes}
\usepackage{subcaption}

\usepackage{algorithm}
\usepackage[noend]{algpseudocode}
\usepackage{colortbl} 
\usepackage{makecell}

\newcommand{\tabitem}{~~\llap{\textbullet}~~}

\definecolor{ruleblue}{RGB}{0, 0, 255}

\newcommand{\sh}[1]{\textcolor{blue}{#1}}
\newcommand{\geoclip}{GeoCLIP}
\newcommand{\geovlm}{GeoDANO}
\newcommand{\geofeat}{geometric premises}
\newcommand{\captionstyle}{GeoCLIP-style}

\usepackage{enumitem} 

\usepackage{pict2e}

\newcommand{\measuredrightanglewithsquare}{%
  \mathord{%
    \mspace{1mu}%
    \text{\msquare}%
    \mspace{1mu}%
  }%
}

\newcommand{\msquare}{%
  \setlength{\unitlength}{1ex}%
  \begin{picture}(1,1)
  \roundcap
  \polyline(0,1)(0,0)(1,0)
  \polyline(0, 0.5)(0.5, 0.5)(0.5,0)
  \end{picture}%
}

\title{GeoDANO: Geometric VLM with Domain Agnostic Vision Encoder}

\author{
 \textbf{Seunghyuk Cho\textsuperscript{1}},
 \textbf{Zhenyue Qin\textsuperscript{3}},
 \textbf{Yang Liu\textsuperscript{3}},
 \textbf{Youngbin Choi\textsuperscript{1}},
\\
 \textbf{Seungbeom Lee\textsuperscript{1}},
 \textbf{Dongwoo Kim\textsuperscript{1,2}}
\\
 \textsuperscript{1}Graduate School of Artificial Intelligence, POSTECH,
\\
 \textsuperscript{2}Department of Computer Science and Engineering, POSTECH, 
\\
 \textsuperscript{3}Australian National University
\\
 \small{
   \textbf{Correspondence to:} Dongwoo Kim \href{mailto:dongwoo.kim@postech.ac.kr}{<dongwoo.kim@postech.ac.kr>}
 }
}

\begin{document}
\maketitle

\begin{abstract}
    We introduce GeoDANO, a geometric vision-language model (VLM) with a domain-agnostic vision encoder, for solving plane geometry problems. Although VLMs have been employed for solving geometry problems, their ability to recognize geometric features remains insufficiently analyzed. 
To address this gap, we propose a benchmark that evaluates the recognition of visual geometric features, including primitives such as dots and lines, and relations such as orthogonality.
Our preliminary study shows that vision encoders often used in general-purpose VLMs, e.g., OpenCLIP, fail to detect these features and struggle to generalize across domains. To overcome the limitation, we develop GeoCLIP, a CLIP-based model trained on synthetic geometric diagram–caption pairs. Benchmark results show that GeoCLIP outperforms existing vision encoders in recognizing geometric features. We then propose our VLM, GeoDANO, which augments GeoCLIP with a domain adaptation strategy for unseen diagram styles. GeoDANO outperforms specialized methods for plane geometry problems and GPT-4o on MathVerse. The implementation is available at \begingroup\hypersetup{urlcolor=magenta}\url{https://github.com/ml-postech/GeoDANO}\endgroup.
\end{abstract}

\section{Introduction}

Large language models (LLMs) have achieved remarkable success in automated math problem solving, particularly through code-generation capabilities integrated with proof assistants~\citep{lean,isabelle,POT,autoformalization,MATH}. Although LLMs excel at generating solution steps and correct answers in algebra and calculus~\citep{math_solving}, their unimodal nature limits performance in plane geometry, where the solution depends on both diagram and text~\citep{math_solving}. 

\begin{figure}
    \centering
    \begin{subfigure}[t]{.49\linewidth}
        \centering
        \includegraphics[width=\linewidth]{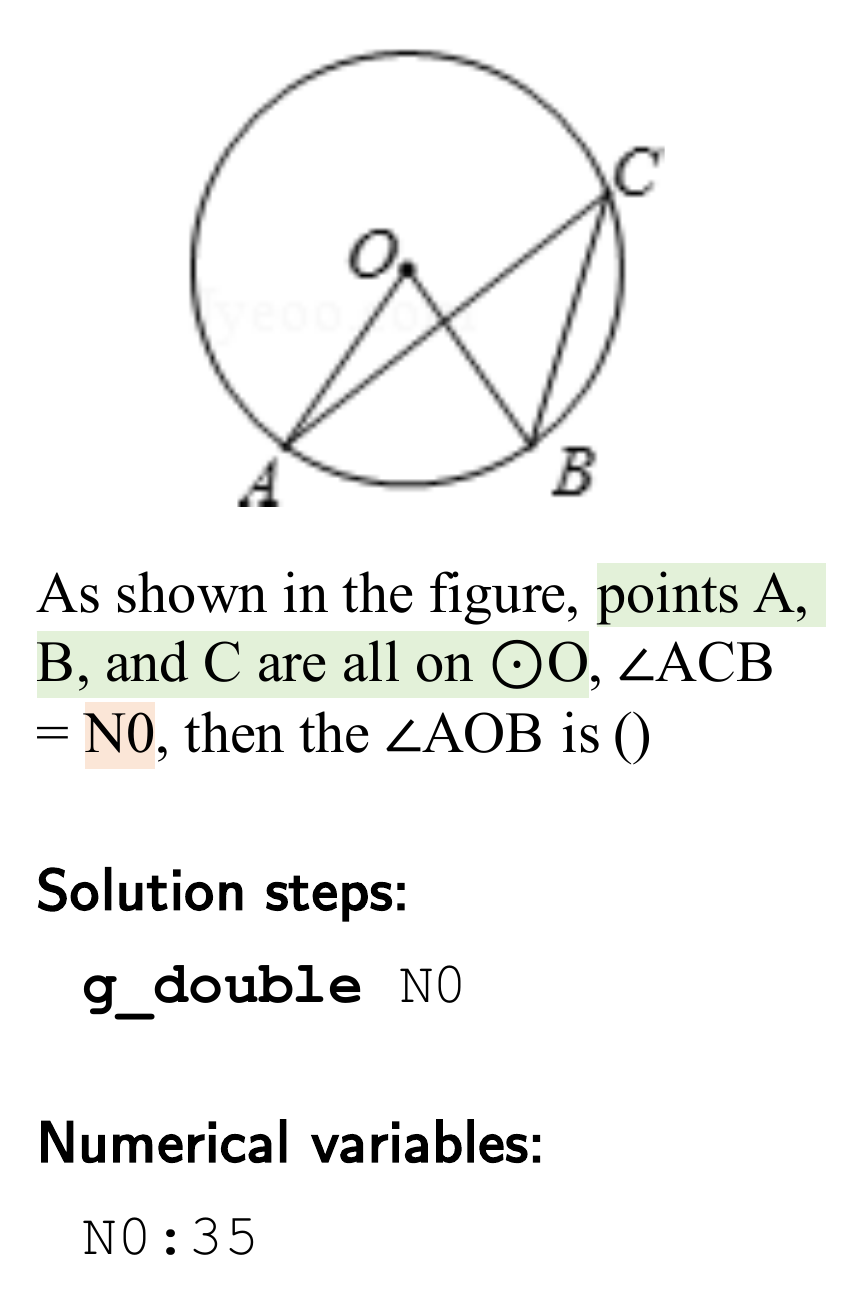}
        \caption{GeoQA}
        \label{fig:geoqa_example}
    \end{subfigure}
    \begin{subfigure}[t]{.48\linewidth}
        \centering
        \includegraphics[width=\linewidth]{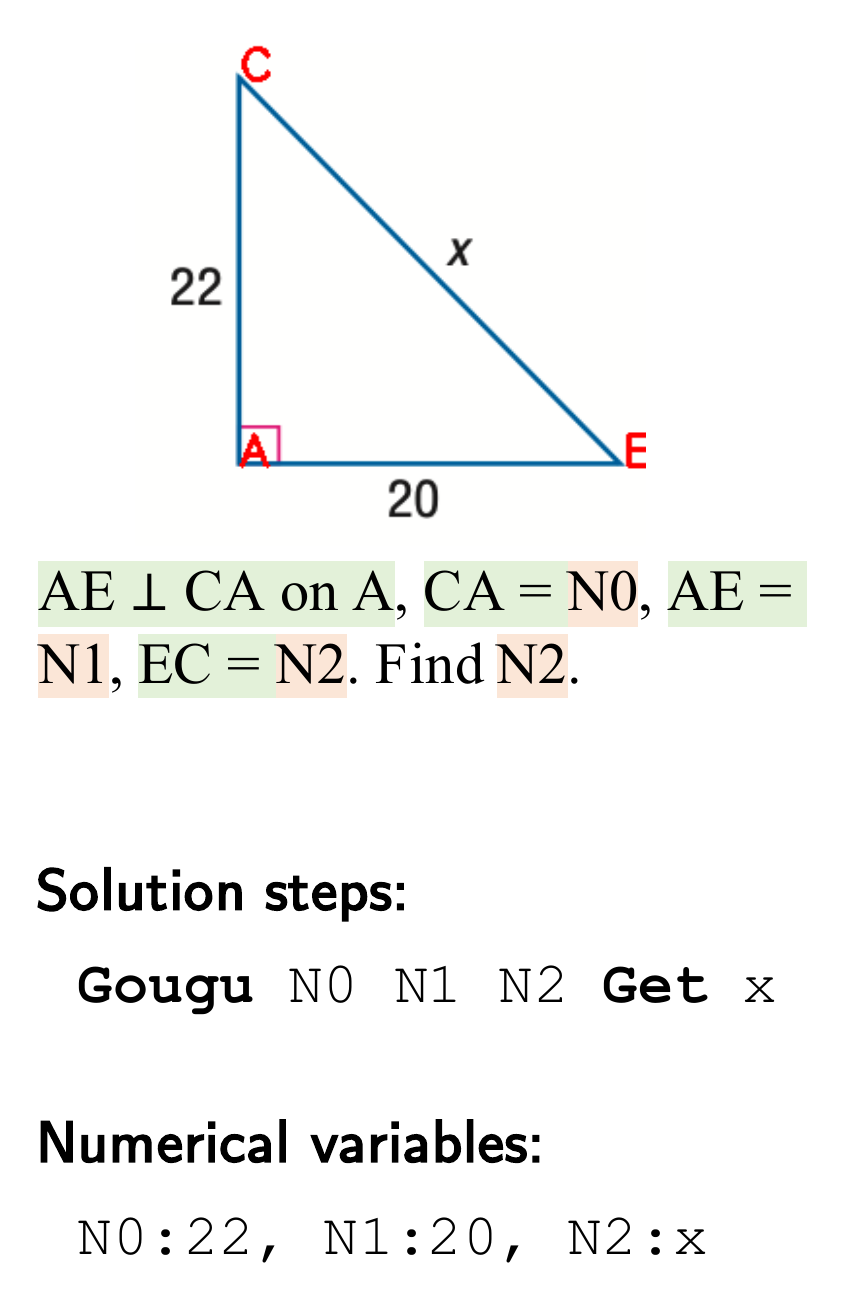}
        \caption{PGPS9K}
        \label{fig:pgps9k_example}
    \end{subfigure}
    \caption{
    Examples of diagram-caption pairs and their solution steps written in formal languages from the GeoQA and PGPS9k datasets. The problem description highlights the visual geometric premises and numerical variables in green and red, respectively. A significant difference in the style of the diagram and formal language can be observed. 
    \label{fig:pgps_examples}
    }
\end{figure}

Specialized vision-language models (VLMs) have accordingly been developed for plane geometry problem solving (PGPS)~\citep{geoqa,unigeo,intergps,pgps,GOLD,LANS,geox}. Yet, whether these models genuinely leverage diagrams or rely almost exclusively on textual features remains unclear. This ambiguity arises because existing PGPS datasets typically embed sufficient geometric details within problem statements, potentially making the vision encoder unnecessary~\citep{GOLD}. \cref{fig:pgps_examples} illustrates example questions from GeoQA and PGPS9K, where solutions can be derived without referencing the diagrams.

We propose a new benchmark created via a synthetic data engine, which systematically evaluates the ability of VLM vision encoders to recognize geometric premises. Our empirical findings reveal that previously suggested self-supervised learning (SSL) approaches, e.g., vector quantized variational auto-encoder (VQ-VAE)~\citep{unimath} and masked auto-encoder (MAE)~\citep{scagps,geox}, and widely adopted encoders, e.g., OpenCLIP~\citep{clip} and DinoV2~\citep{dinov2}, struggle to detect geometric features such as perpendicularity and degrees. 

To this end, we propose \geoclip{}, a model pre-trained on a large corpus of synthetic diagram–caption pairs. By varying diagram styles (e.g., color, font size, resolution, line width), \geoclip{} learns robust geometric representations and outperforms prior SSL-based methods on our benchmark. Building on \geoclip{}, we introduce a few-shot domain adaptation technique that efficiently transfers the recognition ability to real-world diagrams. We finally propose \geovlm{} by combining this domain-adapted GeoCLIP with an LLM, forming a domain-agnostic VLM for solving PGPS tasks. 

In our experiments on MathVerse~\citep{mathverse}, which encompasses diverse plane geometry tasks and diagram styles, \geovlm{} consistently outperforms both task-specific PGPS models and generalist VLMs. 
Ablation studies confirm the effectiveness of our domain adaptation strategy, showing improvements in optical character recognition (OCR)-based tasks and robust diagram embeddings across different styles. 

We summarize the contributions as follows:
We propose a novel benchmark for systematically assessing how well vision encoders recognize geometric premises in plane geometry diagrams~(\cref{sec:visual_feature}); We introduce \geoclip{}, a vision encoder capable of accurately detecting visual geometric premises~(\cref{sec:geoclip}), and a few-shot domain adaptation technique that efficiently transfers this capability across different diagram styles (\cref{sec:domain_adaptation});
We show that our VLM, named \geovlm{}, incorporating domain-adapted GeoCLIP, surpasses existing specialized PGPS VLMs and generalist VLMs on the MathVerse benchmark~(\cref{sec:experiments}) and effectively interprets diverse diagram styles~(\cref{sec:abl}).

\section{Related Work}

In this section, we summarize the known PGPS benchmarks and models. Detailed comparison with previous work is reported in \cref{appendix:related_work}.

\subsection{PGPS benchmarks}
\label{sec:relatedwork_benchmark}


Several studies have introduced benchmarks for PGPS, including a set of diagrams and corresponding problem and solution descriptions~\citep{geoqa,intergps,pgps,unigeo}. The problem and solution descriptions are provided in natural languages or formal languages. Often, the solution steps are provided in the form of formal language. Given the dataset, the goal of PGPS is to train a model that produces a valid solution as an executable program.

Recently, MathVerse~\citep{mathverse} provides an alternative view to existing PGPS benchmarks by directly encoding the geometric properties and relations into the diagrams rather than text description. Therefore, it is impossible to produce a valid solution without recognizing the necessary information from diagrams. 
CogAlign~\citep{cogalign} introduces a benchmark that evaluates the spatial relationship understanding of pretrained vision encoders via linear probing. However, the questions in this benchmark focus purely on spatial relationships and do not involve symbols representing geometric relations.

\subsection{Program generation based PGPS}

A core challenge in program generation-based PGPS is processing both diagrams and text to interpret \geofeat{}. One approach tackles the challenge by converting a diagram into alternative representations such as lists of geometric primitives and relations that can be represented as text~\citep{geos,geos-plus,intergps, GOLD, pgdp, geodrl}. Although reducing the problem to a single modality can be effective, building such converters typically requires labeled diagrams, which are expensive to collect and eventually limit generalization across diverse diagram styles.

Another line of research typically employs vision-language models (VLMs), where a VLM comprises a vision encoder and a language model~\citep{pgps,geoqa,geoqa-plus,scagps,unigeo,unimath,geox,LANS}. 
The vision encoder produces a visual embedding from the diagram, and the language model then generates solution steps in an autoregressive manner, conditioned on the textual description and the visual embedding.
While the VLMs apply to various diagram formats, the visual \geofeat{} perception of the VLMs remains underexplored due to the abundance of textual information in existing benchmarks. 

\subsection{Contrastive learning in PGPS}

Contrastive learning is applied in diverse domains such as computer vision~\citep{facenet} and natural language processing~\citep{simcse}. 
In the context of PGPS, contrastive learning is employed to address domain-specific challenges. GeoX~\citep{geox} applies contrastive learning to the adapter layer of the VLM to enhance formal language comprehension. 
GeoGLIP~\citep{sve-math} utilizes grounded language-image pre-training with synthetic diagram and junction, boundary triples.
Other approaches train the vision encoder itself using the contrastive language-image pre-training (CLIP)~\citep{clip} objective: LANS~\citep{LANS} aligns patch embeddings from a vision Transformer (ViT) with text token embeddings if they describe the same point.
MAVIS~\citep{mavis} employs diagram–caption pairs generated by a synthetic engine for CLIP, where the captions contain all the information in the diagram.

\section{Benchmark for Geometric Premises}
\label{sec:visual_feature}

In this section, we first develop a benchmark for evaluating a vision encoder's performance in recognizing geometric features from a diagram. We then report the performance of well-known vision encoders on this benchmark.


\subsection{Benchmark preparation}
\label{sec:synthetic_data_engine}

\begin{figure*}[t!]
    \centering
    \newcommand{\textbox}[2]{
        \begin{minipage}[t][3.8cm][b]{\linewidth}
            \raggedright\scriptsize\baselineskip=10pt
            \sffamily
            \textbf{Q:} #1 \\[5pt]
            \textbf{Choices:} \\[0pt] #2
        \end{minipage}
    }
    \begin{subfigure}[t]{.19\linewidth}
        \centering
        \begin{tikzpicture}
            \node[anchor=south west] (img) at (0,2) {\includegraphics[width=.85\linewidth]{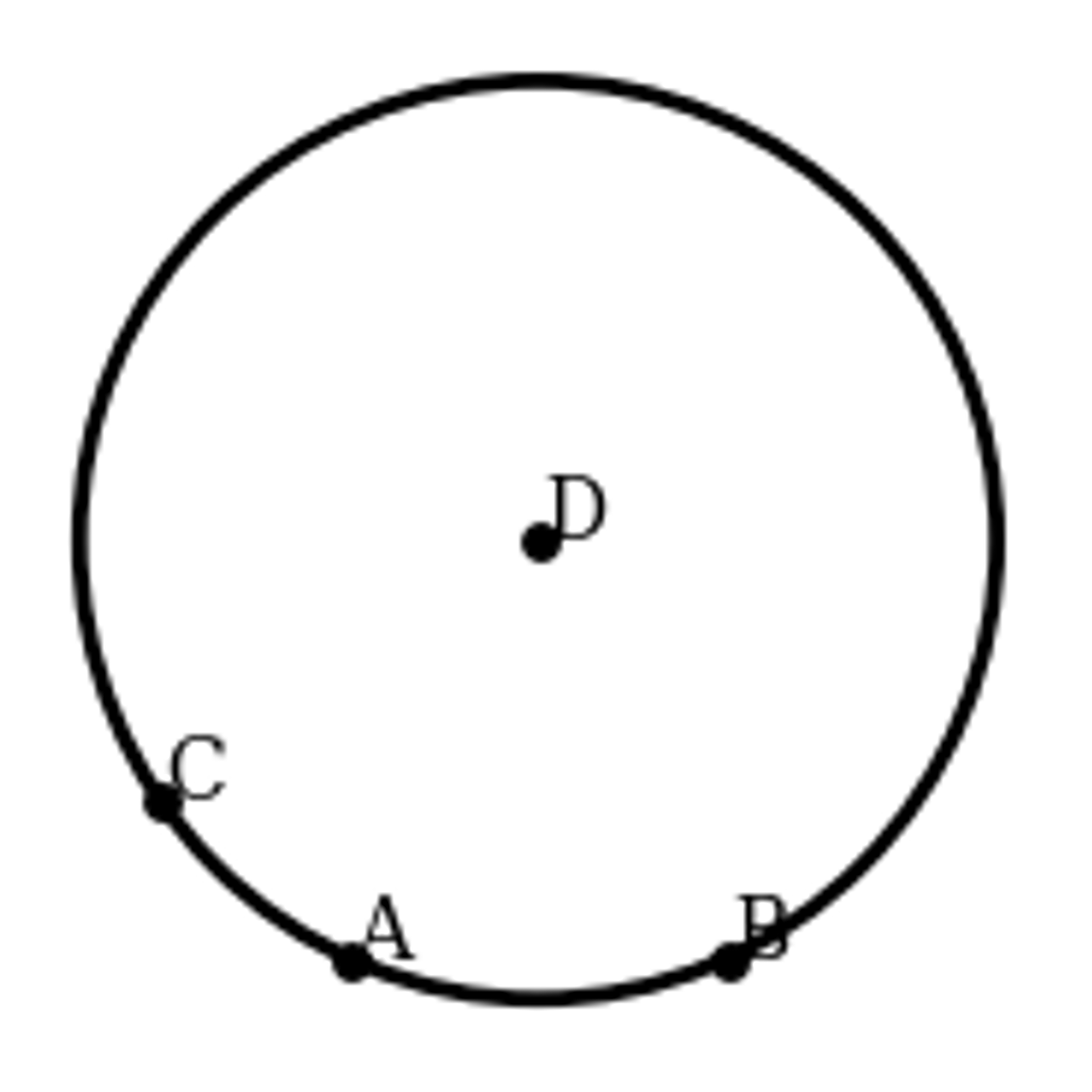}};
            \node[anchor=south west, align=left] at (0, 0) {\textbox{
                How many points are \\[0pt] on the circle?
            }{
                (A) 0 \quad (B) 1 \\[0pt]
                (C) 2 \quad (D) 3 \quad (E) 4
            }};
        \end{tikzpicture}
        \caption{Concyclic}
    \end{subfigure}
    \begin{subfigure}[t]{.19\linewidth}
        \centering
        \begin{tikzpicture}
            \node[anchor=south west] (img) at (0,2) {\includegraphics[width=.85\linewidth]{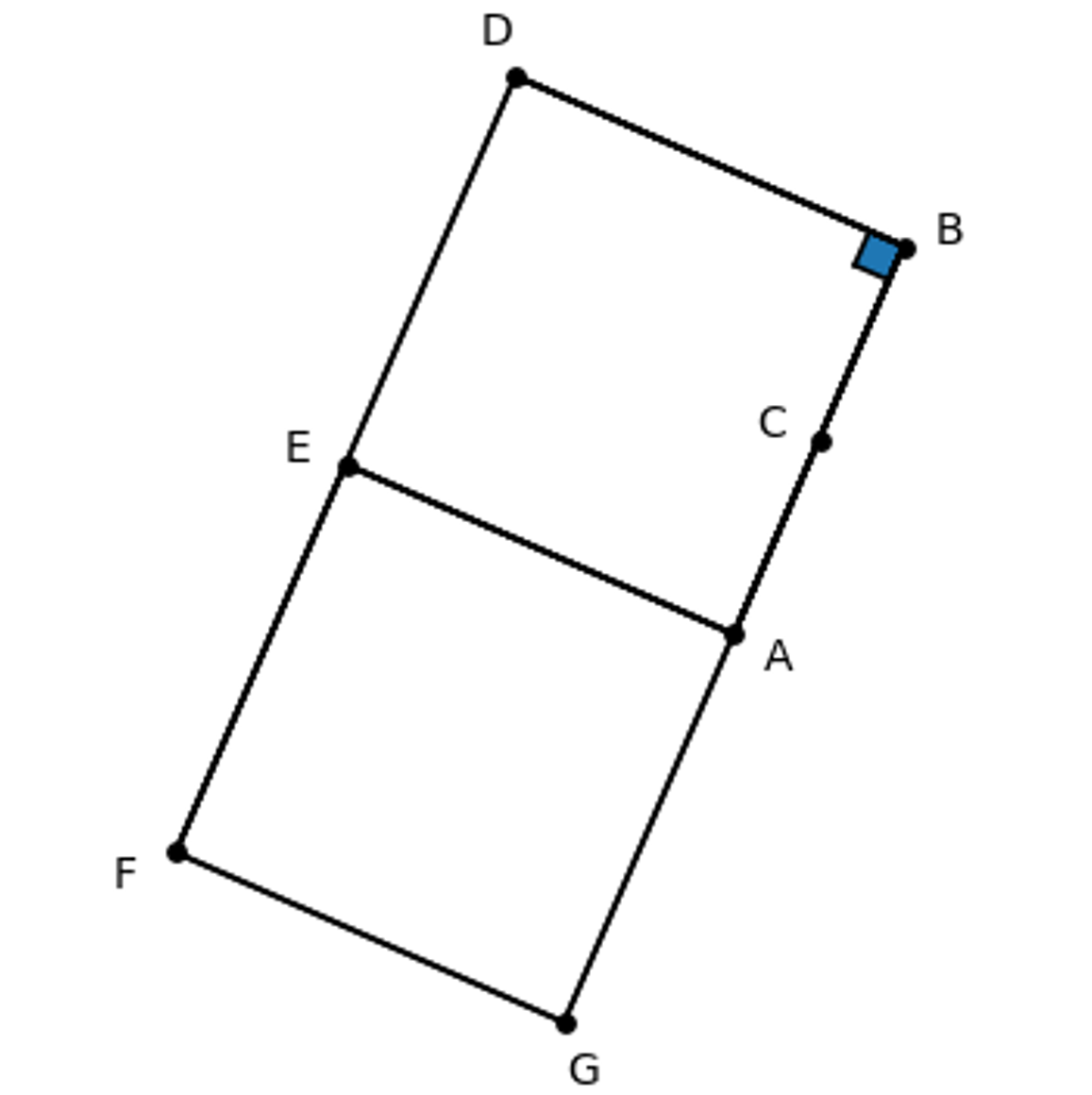}};
            \node[anchor=south west, align=left] at (0, 0) {\textbox{
                How are $\overline{\rm {AB}}$ and $\overline{\rm {BC}}$ related?
            }{
                (A) Perpendicular \\[0pt]
                (B) Collinear (C) Otherwise
            }};
        \end{tikzpicture}
        \caption{TwoLines}
    \end{subfigure}
    \begin{subfigure}[t]{.19\linewidth}
        \centering
        \begin{tikzpicture}
            \node[anchor=south west] (img) at (0,2) {\includegraphics[width=.85\linewidth]{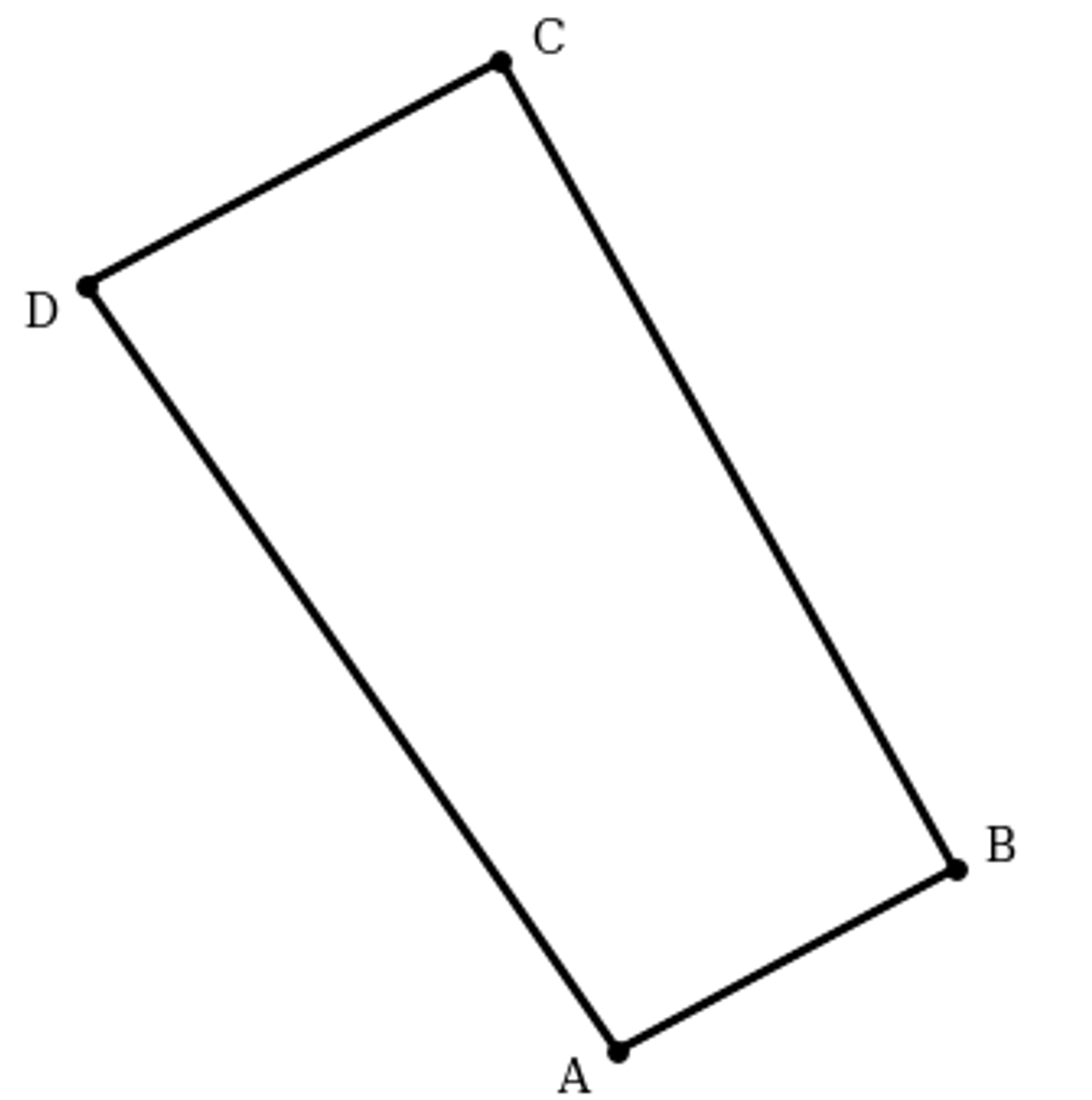}};
           \node[anchor=south west, align=left] at (0, 0){\textbox{
                What kind of object is \\[0pt] in the diagram?
            }{
                (A) Segment (B) Triangle \\[0pt]
                (C) Square (D) Pentagon
            }};
        \end{tikzpicture}
        \caption{ObjectShape}
    \end{subfigure}
    \begin{subfigure}[t]{.19\linewidth}
        \centering
        \begin{tikzpicture}
            \node[anchor=south west] (img) at (0,2) {\includegraphics[width=.85\linewidth]{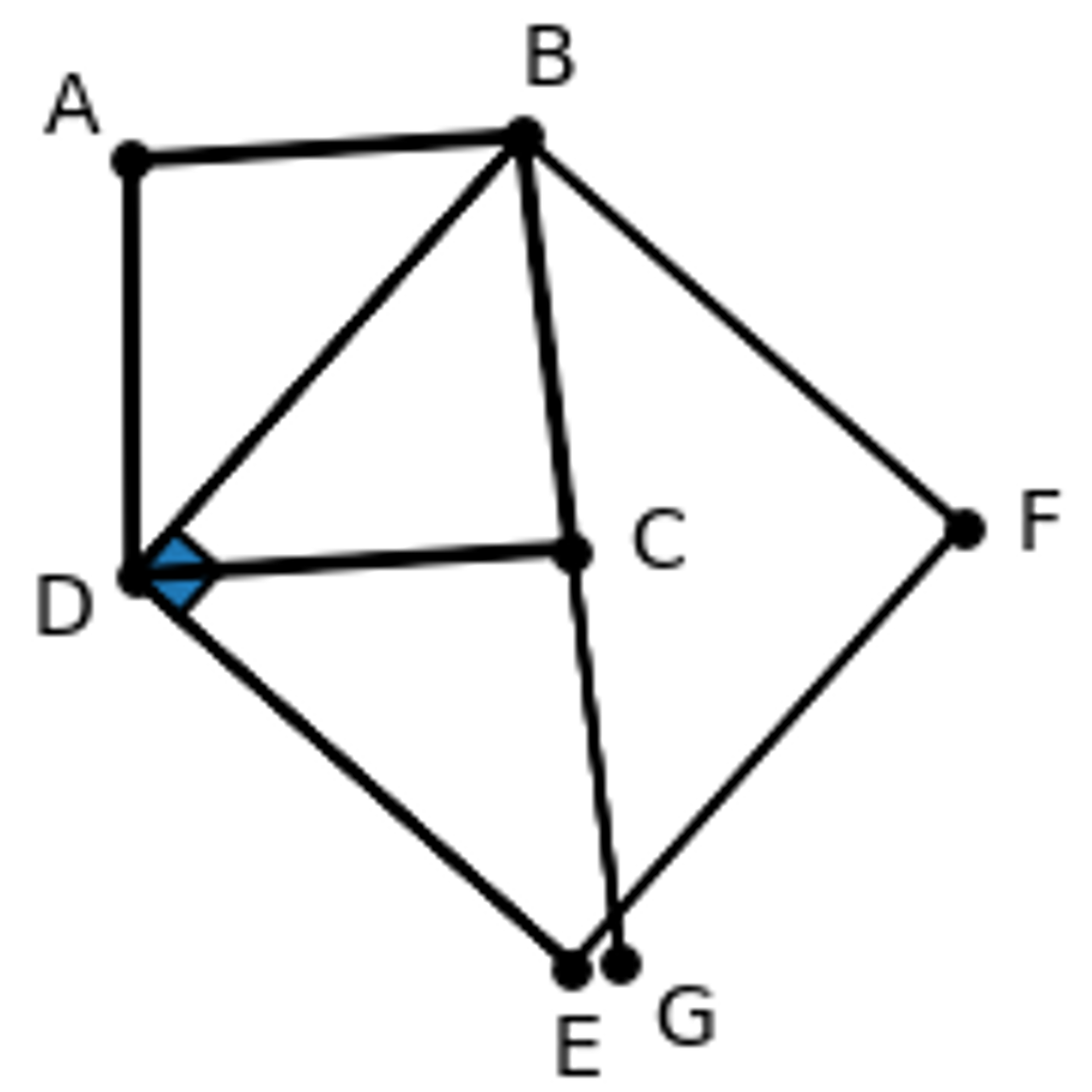}};
            \node[anchor=south west, align=left] at (0, 0) {\textbox{
                What is the shape of $\square \rm{ABCD}$?
            }{
                (A) Parallelogram \\[0pt]
                (B) Trapezoid (C) Rectangle
            }};
        \end{tikzpicture}
        \caption{SquareShape}
    \end{subfigure}
    \begin{subfigure}[t]{.19\linewidth}
        \centering
        \begin{tikzpicture}
            \node[anchor=south west] (img) at (0,2) {\includegraphics[width=.85\linewidth]{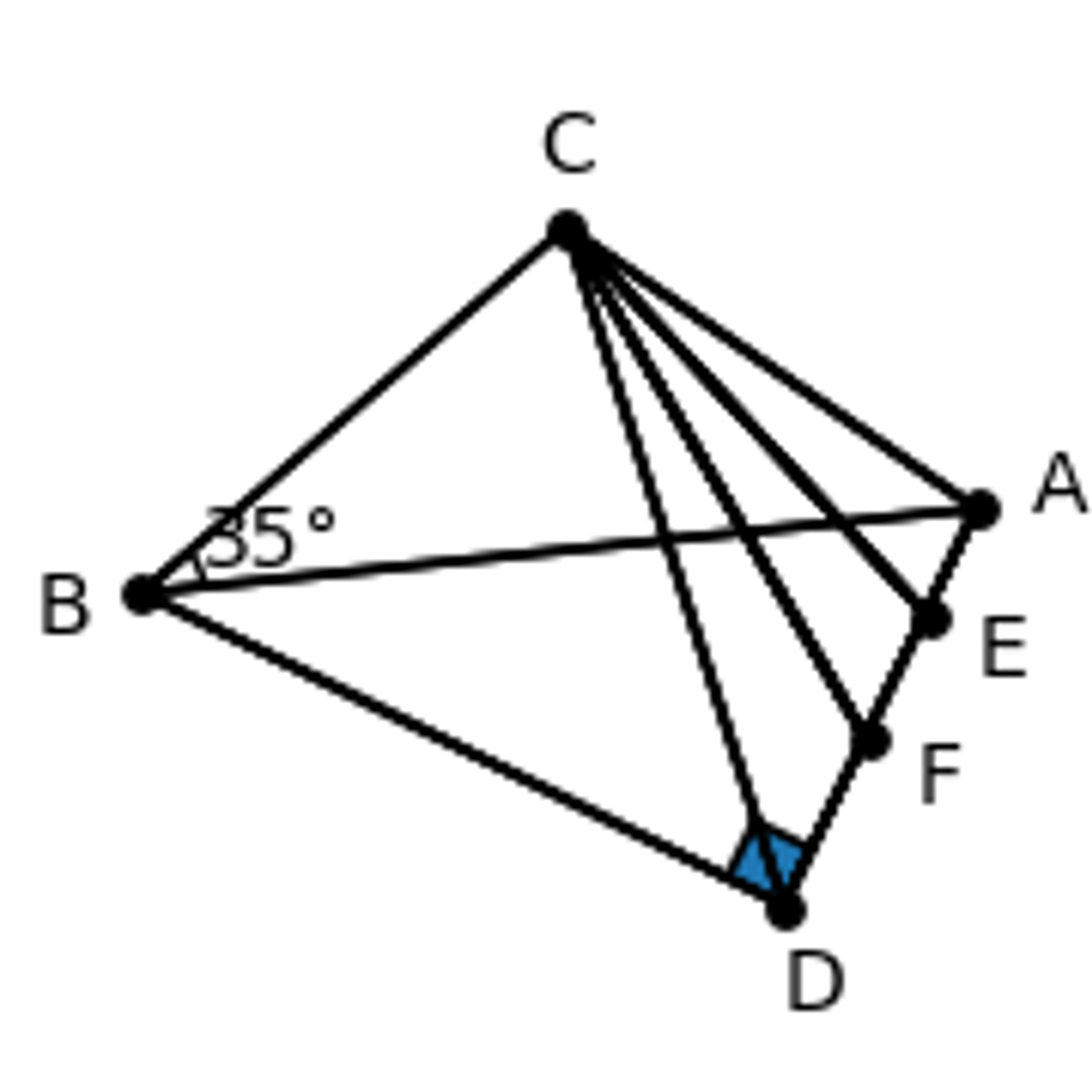}};
            \node[anchor=south west, align=left] at (0, 0) {\textbox{
                What is the degree of $\angle \rm{ABC}$?
            }{
                (A) $15^\circ$ \quad (B) $20^\circ$ \\[0pt]
                (C) $25^\circ$ \quad $\cdots$ \quad (N) $75^\circ$
            }};
        \end{tikzpicture}
        \caption{AngleDetection}
    \end{subfigure}
    \caption{Illustration of the proposed visual feature perception benchmark. We introduce five different diagram classification tasks that require visual feature perception to answer geometry-related questions.}
    \label{fig:benchmark}
\end{figure*}

We design our benchmark as simple classification tasks. By investigating PGPS datasets, we identify that recognizing \emph{geometric primitives}, such as points and lines, and geometric properties representing \emph{relations between primitives}, such as perpendicularity, is important for solving plane geometry problems. Recognized information forms \emph{geometric premises} to solve the problem successfully. To this end, we carefully curate five classification tasks as follows:
\begin{itemize}[leftmargin=*] \setlength\itemsep{1px}
    \item \textbf{Concyclic}: A circle and four points are given. The task is identifying how many of those points lie on the circle.
    \item \textbf{TwoLines}: Two lines, AB and BC, are given alongside other geometric objects. The task is determining whether AB and BC are perpendicular, collinear, or neither.
    \item \textbf{ObjectShape}: A given diagram includes one of the following geometric objects: a segment, triangle, square, or pentagon. The task is to classify which object is present.
    \item \textbf{SquareShape}: A diagram including a square ABCD and other geometric objects is given. The task is to classify whether the square is a trapezoid, parallelogram, or rectangle.
    \item \textbf{AngleDetection}: A diagram is given with at least three points: A, B, and C. The task is to classify the correct angle of ABC from \(\{15^\circ, 20^\circ, \ldots, 75^\circ\}\). 
\end{itemize}
An example of each task is provided in \cref{fig:benchmark}.

Our benchmark is built on top of AlphaGeometry~\citep{alphageometry}, which is designed to solve IMO-style plane geometry problems. The program provides useful functions such as formal language describing plane diagrams. The language predefines a set of geometric premises listed in \cref{tab:alphageometry}, including all necessary properties to define our benchmark tasks. In addition, once a diagram description is given in formal language, the program renders a corresponding diagram with varying fonts, colors, widths, orientations, and resolutions, allowing us to have diagrams with diverse styles. 


We create question-and-answer pairs based on AlphaGeometry.
To sample a diverse set of questions and answers, we first establish a foundational geometric structure corresponding to the key problem of the task using the formal language provided by AlphaGeometry. For instance, in the AngleDetection benchmark, we specify AlphaGeometry problems that guarantee the presence of angle ABC with degree between 15° and 75° in the diagram.

We then execute \cref{alg:sampling} to generate diagrams and corresponding answers from these predefined geometric specifications. 
Specifically, \cref{alg:sampling} first samples an initial AlphaGeometry problem containing the essential geometric premise, then incrementally adds random geometric primitives and relations to diversify the diagrams. 
Finally, the answer to the generated diagram is determined from the initial sampled AlphaGeometry problem. 
Importantly, in the TwoLines, SquareShape, and AngleDetection benchmarks, the answer obtained from the initial AlphaGeometry problem remains unchanged and visually present despite adding other geometric elements. 
In contrast, for ObjectShape and Concyclic benchmarks, no additional geometric elements are introduced, further ensuring the accuracy of the answer. 
Consequently, the answers derived from these formal specifications consistently match the generated diagrams.

For each task, we generate 50,000, 10,000, and 10,000 question-and-answer pairs for training, validation, and testing, respectively.
Additional details on the benchmark generation process are available at \cref{sec:role_textual_description}.

\subsection{Results}
\label{sec:benchmakr_results}
\begin{table}[t!]
    \centering
    \resizebox{\linewidth}{!}{
    \begin{tabular}{l l c c c c c }
        \toprule
        & Models & \begin{tabular}{@{}c@{}}Object \\ Shape\end{tabular} & \begin{tabular}{@{}c@{}}Con \\ cyclic\end{tabular} & \begin{tabular}{@{}c@{}}Two \\ Lines\end{tabular} & \begin{tabular}{@{}c@{}}Square \\ Shape\end{tabular} & \begin{tabular}{@{}c@{}}Angle \\ Detection\end{tabular} \\
        \midrule
        \multirow{8}{*}{%
    \rotatebox[origin=c]{90}{%
        \parbox{1.3cm}{\centering \footnotesize \emph{Baseline}}%
    }%
}  
        &OpenCLIP & \textbf{100.00} & 99.13 & 86.57 & 85.20 & 64.81 \\
        &SigLIP & \textbf{100.00} & \textbf{99.71} & 89.26 & 89.31 & 76.86 \\
        &DinoV2 & \textbf{100.00} & 98.01 & 85.30 & 91.24 & 22.43 \\
        &ConvNeXT & \textbf{100.00} & 99.20 & 89.39 & 88.13 & 61.84 \\
        &MAVIS-CLIP & 91.64 & 71.78 & 58.84 & 60.48 & 16.12\\
        &GeoGLIP & 98.40 & 91.97 & 60.22 & 63.10 & 11.58\\
        &AutoGeo & 99.85 & 91.48 & 78.90 & 90.40 & 23.24 \\
        &FM-ViT & 96.89 & 87.29 & 61.73 & 64.25 & 15.90 \\
        \midrule
        \multirow{3}{*}{%
    \rotatebox[origin=c]{90}{%
        \parbox{1.3cm}{\centering \footnotesize \emph{SSL}}%
    }%
}  
        &Jigsaw & 86.11 & 63.85 & 49.98 & 61.88 & 11.44 \\
        &MAE & 93.99 & 72.25 & 71.73 & 82.70 & 13.08 \\
        &VQ-VAE & 63.05 & 60.97 & 48.10 & 57.35 & 9.22 \\
        \midrule
        \multirow{3}{*}{%
    \rotatebox[origin=c]{90}{%
        \parbox{1.3cm}{\centering \footnotesize \emph{GeoCLIP}}%
    }%
}  
        &GeoCLIP (F $\times$) & 99.52 & 98.61 & 88.33 & 86.76 & 65.68 \\
        &GeoCLIP (2K) & 99.32 & 98.73 & 94.73 & 89.22 & 74.95 \\
        &GeoCLIP & 99.21 & 99.24 & \textbf{96.05} & \textbf{95.95} & \textbf{78.56} \\
        \bottomrule
    \end{tabular}
    }
    \caption{Results on the proposed visual feature benchmark. We report the test accuracy of the models with the best validation performance. }
    \label{tab:linear_probing}
\end{table}

We evaluate four eight adopted vision encoders for the open-sourced VLMs: OpenCLIP~\citep{clip}, SigLIP~\citep{siglip}, DinoV2~\citep{dinov2}, ConvNeXT~\citep{convnext}, GeoGLIP~\citep{sve-math}, FM-ViT~\citep{fm-vit}, AutoGeo~\citep{autogeo}, and MAVIS-CLIP~\citep{mavis}.
For MAVIS-CLIP, FM-ViT, and AutoGeo, we train OpenCLIP on different datasets under their respective configurations: MAVIS-CLIP is trained on MAVIS-Caption following the MAVIS setup, AutoGeo is trained on synthetic diagram–caption pairs generated by AutoGeo using the MAVIS setup, and FM-ViT is trained on Geo170K~\citep{gllava} and MAVIS-Caption~\citep{mavis} following the FM-ViT setup.


To evaluate the vision encoder, we use linear probing, i.e., adding a linear layer on top of each encoder as a prediction head and training the linear layer from scratch while freezing the parameters of the vision encoder. We use a training set to train the prediction head and report the test accuracy with the best validation performance. The details for the hyper-parameters are described in \cref{sec:hparams}.

As shown in \cref{tab:linear_probing}, many existing vision encoders relatively well recognize the shape of objects but fail at the correct angle between two lines. The encoders also show some difficulties in recognizing the shape of a square and the relationship between two lines. Although the result may seem satisfactory at a glance, these errors will propagate to the downstream tasks when combined with LLMs.

\section{\geoclip: Enhanced Vision Encoder}

In this section, we first propose \geoclip{}, a new vision encoder designed to recognize geometric premises from diverse styles of diagrams.
To transfer the recognition to real-world PGPS benchmarks, we then propose a domain adaptation technique for \geoclip{} that leverages a small set of diagram–caption pairs from target domains. 



\subsection{Training \geoclip{}}
\label{sec:geoclip}

We propose a \geoclip{}, a vision encoder trained with the CLIP objective with a newly developed 200,000 diagram-caption examples.
From the random diagram generator developed in \cref{sec:synthetic_data_engine}, we additionally sample 200,000 diagrams written in the formal language. Directly rendering these samples can result in a diagram that may not preserve the geometric properties. For example, the perpendicularity between two lines cannot be observed from the diagram without having the right angle sign, i.e., $\measuredrightanglewithsquare$. Therefore, we ensure to render the images containing all necessary geometric premises from their visual illustration.

For the caption of a diagram, we filter out some geometric properties from the original description of a diagram used to render the image. Specifically, we only keep the following four properties, concyclic, perpendicularity, angle measures, and length measures, from the visual premises shown in \cref{tab:alphageometry}. After that, we convert the remaining descriptions written in the formal language into natural language. We filter out some properties for two reasons.
First, some properties are not recognizable from the rendered diagram without additional information, e.g., congruency. These properties are listed as non-visual premises in \cref{tab:alphageometry}. Second, collinearity and parallelity occur so frequently that they can marginalize others.
Some examples of generated captions after filtering and translating are provided in the right-most column of \cref{fig:alphageometry}. We call the filtered caption as \emph{\captionstyle{} caption}.

With this dataset, we fine-tune OpenCLIP~\citep{clip} via the CLIP objective, formulated as:
\begin{align}
    &\mathcal{L}_{\textrm{CLIP}}(\mathcal{D}, g, h) := \nonumber \\
    &\,\,\,\,\,\mathbb{E}_{\mathcal{D}} \!\biggl[ -\log \frac{\exp \bigl( g(D_i)^T \, h(X_i) / \tau \bigr)}{\sum_{X \in \{X_i\}_i} \exp \bigl( g(D_i)^T \, h(X) / \tau \bigr)} \biggr],
    \label{eq:clip}
\end{align}
where \(\mathcal{D} := \{(D_i, X_i)\}_{i=1}^N\) is the diagram-caption pairs, $g$ is the vision encoder, $h$ is the text encoder, and \(\tau\) is a temperature parameter.
We named the resulting vision encoder as \geoclip{}. \cref{sec:hparams} provides the details, including hyper-parameters.

We compare the performance of \geoclip{} to other self-supervised approaches trained with the same dataset. We test three self-supervised approaches: Jigsaw~\citep{geoqa, geoqa-plus}, MAE~\citep{scagps, geox}, and VQ-VAE~\citep{unimath}, used in previous work to improve the recognition performance of plane diagrams. We use the same architecture used for \geoclip{} for Jigsaw and MAE with the hyper-parameters used in the previous works. For VQ-VAE, we follow the architecture of \citet{unimath}. 
All model performances are measured through the linear probing used in \cref{sec:benchmakr_results}. 

As shown in \cref{tab:linear_probing}, \geoclip{} recognizes geometric features better than existing baselines and self-supervised methods. The self-supervised approaches generally perform poorly for the benchmark, justifying the choice of the objective. We also compare the performance of \geoclip{} against other encoders such as OpenCLIP. Note that although we outperform the other encoders in tasks such as SquareShape and AngleDetection, these results might be \emph{unfair} compared to the existing pretrained vision encoders, since the training set of \geoclip{} is similar to the diagrams in the benchmark.
The t-SNE plots of the embeddings from the vision encoders are illustrated at \cref{fig:tsne_benchmark}.

We further ablate the filtering process in \geoclip{}. To this end, we compare \geoclip{} with its two variants: \emph{GeoCLIP (F $\times$)}, which uses the captions generated without filtering. We also test \emph{GeoCLIP (2K)}, which is trained on only 2,000 pairs, to see the effectiveness of the large-scale dataset.
The results in \cref{tab:linear_probing} imply that both the filtering and the training set size matter in enhancing geometric properties recognition.

\subsection{Domain adaptation of \geoclip{}}
\label{sec:domain_adaptation}

Although \geoclip{} enhances the geometric premises recognition on the benchmark set, the diagram styles in existing PGPS benchmarks differ, necessitating further adaptation. To overcome this challenge, we propose a few-shot domain adaptation method utilizing a few labeled diagrams.

A domain-agnostic vision encoder must match the same diagrams drawn in different styles. To do so, we need a target domain diagram translated into the source domain style or the source diagrams translated into the target domain style. With these translated images, we can guide the model to focus on key geometric information instead of irrelevant attributes, such as color and font family. However, in practice, it is difficult to obtain the same diagrams with different styles. 

\begin{figure*}
    \centering
    \includegraphics[width=.9\linewidth]{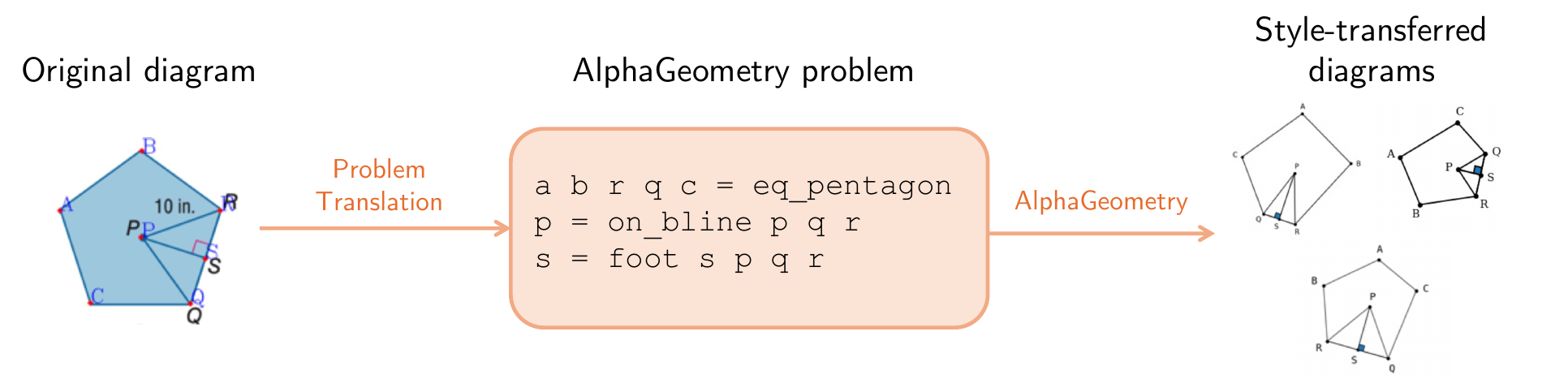}
    \caption{Illustration of the translation process for domain adaptation. We first translate the given geometric diagram from the target domain into an AlphaGeometry problem. 
    We then generate multiple diagrams sharing the same visual geometric premises in the AlphaGeometry style.
    }
    \label{fig:translation_process}
\end{figure*}

We develop a way to translate the target diagrams into the source style.
Thankfully, since well-known PGPS datasets come with diagram captions written in formal languages~\citep{intergps}, we can easily convert them to the AlphaGeometry-style descriptions. 
Given the translated descriptions, we utilize the rendering engine of AlphaGeometry to translate the target domain images into the source domain. 
With the translation, we can generate the same diagram in the source domain style. 
\cref{fig:domain_adaptation_samples} provides examples of the diagram pairs with different styles. However, in some cases, the original description contains geometric premises that are unrecognizable from the diagram, such as \(\angle ACB = 35.0\) in \cref{fig:geoqa_example}. Therefore, we apply the same filtering process used in \geoclip{} to translate the AlphaGeometry-style descriptions into natural languages.
Additional details in the translation process is described in \cref{sec:translation_details} and \cref{fig:translation_process}.


Formally, let $\mathcal{D}_{S} := \{(D_S^{(i)}, X_S^{(i)}) \}_{i=1}^{N_S}$ be the diagram-caption pairs from source domain $S$, e.g., the synthetic diagrams, and let $\mathcal{D}_{T_j} := \{(D_{T_j}^{(i)}, X_{T_j}^{(i)}) \}_{i=1}^{N_{T_j}}$ be the set of diagram-caption pairs of target domain $T_j$, e.g., the PGPS benchmarks. With the translation process described above, we can synthesize a style-transferred diagram-caption pair $(\hat{D}_{T_j}^{(i)}, \hat{X}_{T_j}^{(i)})$ for each diagram $D_{T_j}^{(i)}$ and caption $X_{T_j}^{(i)}$ in target domain $T_j$.

We adapt the domain by fine-tuning the vision encoder through the style-transferred diagram-caption pairs.
Let $\hat{\mathcal{D}}_{T_j}$ be a collection of the original diagram and style-transferred captions, i.e., $\hat{\mathcal{D}}_{T_j} = \{(D_{T_j}^{(i)}, \hat{X}_{T_j}^{(i)})\}_{i=1}^{N_{T_j}}$, and let $\hat{\mathcal{D}}_{T_jS}$ be a collection of the original and style transferred diagram pairs, i.e., $\hat{\mathcal{D}}_{T_jS}= \{(D_{T_j}^{(i)}, \hat{D}_{T_j}^{(i)}) \}_{i=1}^{N_{T_j}}$. The cross-domain adaptation objective is written as
\begin{align}
    &\mathcal{L}_{\textrm{CLIP-DA}}(\mathcal{D}_S, \{\mathcal{D}_{T_j}\}_j, g, h) := 
    \mathcal{L}_{\textrm{CLIP}}(\mathcal{D}_S, g, h) + \nonumber\\
    &\,\,\,\,\,\,\,\,\Sigma_j \mathcal{L}_{\textrm{CLIP}}(\hat{\mathcal{D}}_{T_j}, g, h) + \mathcal{L}_{\textrm{CLIP}}(\hat{\mathcal{D}}_{T_jS}, g, g),
    \label{eqn:da}
\end{align}
where $g$ and $h$ are the vision and text encoders of \geoclip{}, respectively. Note that we do not use the original captions from the target domain, since our goal is to adapt the vision encoder to the target domain, not the text encoder.
\section{Experiments}

In this section, we evaluate the PGPS performance of our VLM equipped with the domain-adapted \geoclip{} on MathVerse~\citep{mathverse}.
We compare its performance against established PGPS baselines. We also present ablation studies highlighting our VLM’s strong visual feature recognition and resilience to domain shifts, both of which are facilitated by the adapted vision encoder.

\begin{figure*}
    \centering
    \begin{subfigure}[t]{.4\linewidth}
        \centering
        \includegraphics[width=\linewidth]{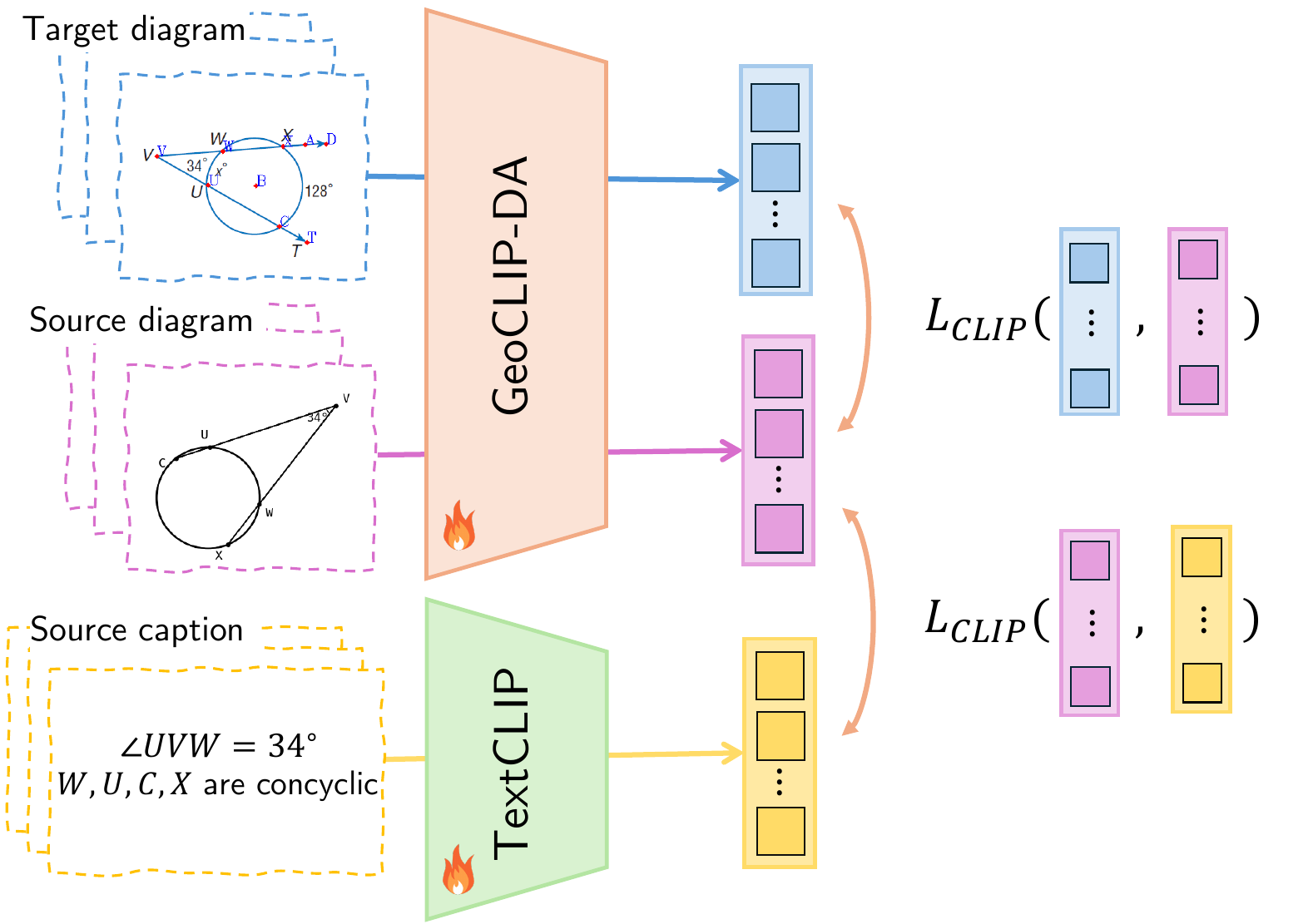}
        \caption{GeoCLIP-DA}
        \label{fig:geoclip-da}
    \end{subfigure}
    \begin{subfigure}[t]{.59
    \linewidth}
        \centering
        \includegraphics[width=\linewidth]{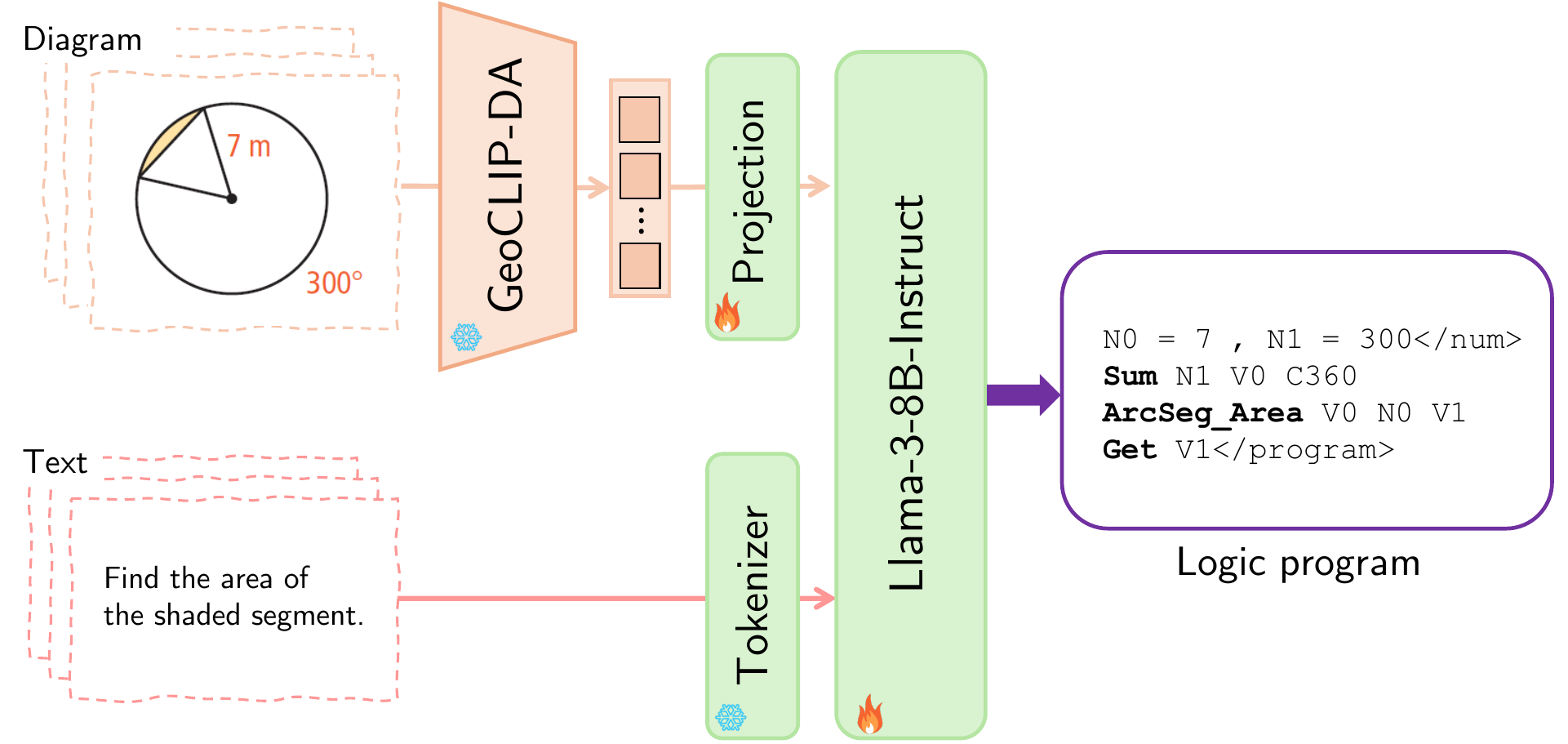}
        \caption{GeoDANO}
        \label{fig:geodano}
    \end{subfigure}
    \caption{The overall training process of GeoDANO. The GeoDANO training consists of two stages. (a) Using the CLIP objective, we first train GeoCLIP-DA by optimizing OpenCLIP's vision and text encoders on synthetic diagram-caption pairs and apply domain adaptation. (b) We then construct GeoDANO by combining GeoCLIP-DA with a projection layer and a language model. We train the projection layer and language model jointly in an end-to-end manner using diagram-text pairs annotated with logic programs.}
    \label{fig:geodano_pipeline}
\end{figure*}

\subsection{Experimental settings and training details}


\paragraph{Datasets.}
We use MathVerse~\citep{mathverse} to measure the performance of VLMs.
MathVerse is a benchmark designed to evaluate both the reasoning and visual-feature recognition capabilities of VLMs, covering plane geometry, solid geometry, and function problems. It is constructed by compiling problems from various sources, including Geometry3K~\citep{intergps}, GeoQA~\citep{geoqa}, and GEOS~\citep{geos}. Each problem is presented in five variants: \emph{text-dominant}, which provides all essential textual information for solving the problem; \emph{text-lite}, which omits descriptive details, e.g., object shapes, from the text; \emph{vision-intensive}, which removes certain textual conditions that can be inferred from remaining information; \emph{vision-dominant}, which relocates numerical measurements, such as angles and lengths, from the text to the diagram; and \emph{vision-only}, which offers only the diagram as input, embedding all text within the diagram. In the following experiments, we focus on plane geometry problems and exclude the vision-only task.


\paragraph{Training details.}
We describe the construction of our \textbf{geo}metric VLM with \textbf{d}omain-\textbf{a}gnostic visio\textbf{n} enc\textbf{o}der, named \geovlm{}. 
Based on \geoclip{} developed in \cref{sec:geoclip}, we apply the domain adaptation to GeoQA and Geometry3K datasets. For the domain adaptation, we randomly sample 50 diagrams and translate the diagram and caption styles following the procedure described in \cref{sec:domain_adaptation}. Finally, \geoclip{} is fine-tuned via \cref{eqn:da}.
We name the GeoQA and Geometry3K adapted \geoclip{} as \geoclip{}-DA.

We combine LLama-3-8b-Instruct~\citep{llama} and \geoclip{}-DA to construct a VLM. The combined model is then fine-tuned again with the training set of GeoQA and PGPS9K to predict the solution program. For PGPS9K, we use the Geometry3K split. Additional details about the training can be found in \cref{sec:vlm_details}.

\paragraph{Modification of training data.}
While previous works focusing on PGPS do not consider optical character recognition (OCR) from diagrams since the benchmark datasets, GeoQA and PGPS9K, provide necessary details in problem descriptions, numerical values can appear within diagrams in real-world settings.
Specifically, an interesting observation from GeoQA and PGPS9K datasets is that the numerical measurements, such as angles, lengths, and volumes, are not written in the problem description but given as additional conditions, and the numerals are substituted as a variable in the problem description as shown in \cref{fig:geoqa_example}. Therefore, the VLM only needs to produce the solution program without having optical character recognition (OCR) from the diagram. The variables are automatically substituted with the actual numbers when the program is executed. Therefore, the vision encoders do not need to learn OCR from the image.

However, this approach cannot be generalized to a broader class of problems where the numerals are embedded in the diagram instead of being written in the problem description. Some variants of MathVerse, such as the vision-dominant problems, fall into this category as well. To incorporate OCR into the solution of the problem, we modify some problem statements in the training set, such that the numerical measurements are only shown in the diagram and not in the statements. We further modify the solution problem so that the solution contains OCR results as a part of the final output. Finally, we unify the language of the solution programs used in GeoQA and PGPS9K by converting GeoQA programs into PGPS9K format. The unification makes the output of VLM consistent since both datasets use different types of formal languages.
\cref{fig:training_data} shows examples of the modified input pairs and solutions, where the first problem statement does not have numerical measurements, and the OCR results are in the part of the output solution program.

\begin{table*}[t!]
    \centering
    \resizebox{\linewidth}{!}{
    \begin{tabular}{l rr c rr c rr c rr }
        \toprule
        \multirow{2}{*}{Models} & \multicolumn{2}{c}{Text Dominant} && \multicolumn{2}{c}{Text Lite} && \multicolumn{2}{c}{Vision Intensive} && \multicolumn{2}{c}{Vision Dominant} \\
        \cmidrule{2-3} \cmidrule{5-6} \cmidrule{8-9} \cmidrule{11-12}
        & Completion $\uparrow$ & Top-10 $\uparrow$ && Completion $\uparrow$ & Top-10 $\uparrow$ && Completion $\uparrow$ & Top-10 $\uparrow$ && Completion $\uparrow$ & Top-10 $\uparrow$  \\ 
        \midrule
        PGPSNet & 4.37 & 14.55 && 2.08 & 12.06 && 2.08 & 11.02 && \multicolumn{1}{c}{-} & \multicolumn{1}{c}{-} \\
        NGS & 6.45 & 34.57 && 6.64 & 28.52 && 5.86 & 26.37 && \multicolumn{1}{c}{-} & \multicolumn{1}{c}{-} \\
        SCA-GPS & 6.84 & 18.16 && 5.66 & 16.80 && 3.52 & 15.23 && \multicolumn{1}{c}{-} & \multicolumn{1}{c}{-} \\
        GeoFormer & 16.22 & 32.85 && 16.84 & 30.77 && 13.10 & 29.11 && \multicolumn{1}{c}{-} & \multicolumn{1}{c}{-} \\
        UniMath-Flan-T5 & 17.88 & 32.43 && 16.42 & 30.56 && 13.93 & 28.27 && \multicolumn{1}{c}{-} & \multicolumn{1}{c}{-} \\
        GeoX-Geo3K & 5.41 & 9.98 && 4.16 & 6.86 && 3.53 & 5.61 && \multicolumn{1}{c}{-} & \multicolumn{1}{c}{-} \\
        GeoX-GeoQA & \textbf{24.32} & 37.42 && 17.26 & 32.43 && 13.51 & 16.25 && \multicolumn{1}{c}{-} & \multicolumn{1}{c}{-} \\
        MAVIS & 21.83 & 42.62 & & 15.80 & 39.50 & & 12.68 & 35.55 & & 3.54 & 10.83 \\
        
        \midrule
        \geovlm{} (OC) & 19.13 & 40.12 && 16.63 & 34.72 && 13.31 & 31.81 && 1.25 & 8.12 \\ 
        \geovlm{} (GC) & 20.37 & 41.79 && 18.09 & 38.25 && 15.80 & 35.34 && 5.62 & 19.38 \\
        \geovlm{} (GC-D) & 22.66 & 43.45 && 21.00 & 38.46 && \textbf{18.30} & 35.76 && 6.67 & 20.42 \\
        \geovlm{} & 23.70 & \textbf{47.82} && \textbf{21.21} & \textbf{45.11} && 18.09 & \textbf{42.20} && \textbf{12.08} & \textbf{36.04} \\
        \bottomrule
    \end{tabular}
    }
    \caption{PGPS accuracy on MathVerse benchmark. We compare the performance of \geovlm{} against PGPS specialist models, which generate a solution program as an output. \geovlm{}-OC, -GC, and -GCD are three variants of our model with different encoders. Further details about these variants can be found in \cref{sec:abl}.}
    \label{tab:mathverse}
\end{table*}
\paragraph{Baselines.}
We use two different types of baseline models for the experiments: PGPS \emph{specialist VLMs} and \emph{generalist VLMs}. Specialist VLMs produce a solution program as an output of a given problem, and generalist VLMs produce a natural language solution as an output.

For the specialist VLMs, we test PGPSNet~\citep{pgps}, NGS~\citep{geoqa}, SCA-GPS~\citep{scagps}, GeoFormer~\citep{unigeo}, UniMath-Flan-T5~\citep{unimath}, GeoX~\citep{geox}, and MAVIS~\citep{mavis}. For GeoX, we use the two variants GeoX-Geo3K and GeoX-GeoQA, which are fine-tuned on Geometry3K and GeoQA, respectively. We mimic MAVIS by replacing the vision encoder of GeoDANO with MAVIS-CLIP, while keeping other components, e.g., the projection layer architecture, language model, and training process, unchanged\footnote{Reproducing MAVIS is not feasible due to the unavailability of both the trained model checkpoint and the complete dataset used for training.}.

For the generalist VLMs, we test two GPT-4o variants~\citep{gpt4o}: gpt-4o-2024-11-20 and gpt-4o-mini-2024-07-18, the InternVL2.5 variants: 8B and 26B models~\citep{internvl2_5}, SPHINX-MoE~\citep{sphinx}, and Math-PUMA-DeepSeek-Math-VL-7B~\citep{math-puma}.

\paragraph{Evaluation metric.}
For each plane geometry problem, both the specialist VLMs and \geovlm{} generate 10 outputs via beam search. Following \citet{pgps}, we then use completion accuracy and top-10 accuracy as our primary evaluation metrics. The completion accuracy assesses whether the first successfully executed solution from the beam is correct; the solutions are reviewed in beam order, and success is recorded if the first executable solution produces the correct answer. Top-10 accuracy examines all ten beam outputs, counting a success if any of these solutions yield the correct result upon execution. Note that, as described before, the specialist VLMs do not have OCR capability. For the evaluation, we feed the correct values to the outputs of these models by using the parser developed in \citet{pgps}. For the models that are trained in Chinese, i.e., NGS and SCA-GPS, we use problem descriptions translated by GPT-4o~\citep{gpt4o}.

To measure the performance of the generalist VLMs, we use multiple-choice questions instead of open-ended questions due to the difficulty in parsing the final answer from free-form text. We use the multiple-choice question provided in MathVerse as an additional input to each problem. We ask VLMs to produce the answer in a pre-specified form. We report the top-1 accuracy of these models.
To compare \geovlm{} against the generalist models, we use the same protocol used in \citet{pgps} to measure the accuracy.

\subsection{Results}
\label{sec:experiments}

\paragraph{Performance against specialist VLMs.}
In \cref{tab:mathverse}, \geovlm{} shows the best performance in almost all the problem variants and metrics except the text-dominant task. Note that the specialist models cannot solve the vision-dominant problems since these problems do not contain variables representing numerical values, such as a length, in the problem description.
When comparing the performance between text and vision-dominant tasks, the top-10 accuracy of \geovlm{} on the vision-dominant task is higher than the top-10 accuracy of the specialist models on the text-dominant task, except for GeoX-GeoQA. Given that the two tasks use the same problem set, the result implies that \geovlm{} performs better than the specialist models without having the geometric premises in the problem description. In other words, our vision encoder can extract geometric premises accurately from the visual information.

\begin{table}[t!]
    \centering 
    \resizebox{\linewidth}{!}{
    \begin{tabular}{l c c c c}
        \toprule
        & \begin{tabular}{@{}c@{}}Text \\ Dominant\end{tabular} & \begin{tabular}{@{}c@{}}Text \\ Lite\end{tabular} & \begin{tabular}{@{}c@{}}Vision \\ Intensive\end{tabular} & \begin{tabular}{@{}c@{}}Vision \\ Dominant\end{tabular} \\
        \midrule
        GPT-4o & 40.35 & 39.18 & 38.01 & 36.95 \\
        GPT-4o-mini & 41.12 & 39.53 & 35.59 & 30.50 \\
        InternVL2.5-8B & 38.30 & 36.26 & 35.09 & 21.99 \\
        InternVL2.5-26B & 42.40 & 40.06 & 38.01 & 38.71 \\
        SPHINX-MoE & 27.49 & 25.15 & 26.61 & 22.58\\
        Math-PUMA & 33.04 & 29.53 & 28.36 & 21.11\\
        GeoX-GeoQA & \textbf{52.05} & 45.91 & 37.43 & - \\
        \geovlm{} & 48.54 & \textbf{49.71} & \textbf{41.81} & \textbf{39.30} \\
        \bottomrule
    \end{tabular}
    }
    \caption{Comparison between \geovlm{} and generalist VLMs on multiple choice questions. }
    \label{tab:mathverse_llm_transposed}
    \vskip -0.19in
\end{table}
\paragraph{Performance against generalist VLMs.}
\cref{tab:mathverse_llm_transposed} reports the performance of generalist VLMs and \geovlm{} on multiple choice questions. \geovlm{} outperforms proprietary closed models, i.e., GPT-4o variants, and open-sourced models, i.e., the InternVL2.5 variants. 
Especially, the performance gap between GeoDANO and InternVL2.5-26B reflects the parameter efficiency of our VLM.
While GeoDANO shows impressive results among the variants, the performance of GeoX-GeoQA degrades dramatically as the visual information moves from the text to the diagram.
Our work is the first to show that the specialist can compete with the generalist in MathVerse.

\subsection{Ablation studies}
\label{sec:abl}

\paragraph{Variation of \geoclip{}.}
We perform a detailed empirical analysis to evaluate how effectively the \captionstyle{} captions and the proposed domain adaptation technique improve \geovlm{}’s performance. Specifically, we compare \geovlm{} against other VLMs trained on the \geoclip{} variants, including OpenCLIP~\citep{clip} and the GeoCLIP without domain adaptation.
We also test a variant of \geoclip{} trained with additional diagram-caption pairs from the target domains without having any filtering process. In this case, we utilize all the data in the training sets.

We show the experimental result in \cref{tab:mathverse}. 
\geovlm{}-OC and \geovlm{}-GC represent the VLM with OpenCLIP and GeoCLIP without domain adaptation, respectively. \geovlm{}-GCD represents the GeoCLIP with additional unfiltered domain captions. \geovlm{} outperforms other variants on most tasks, except the completion accuracy on the vision-intensive task.

\paragraph{OCR performance.}
We assess the accuracy of \geovlm{} and its variants in OCR on the MathVerse diagrams, focusing on the vision-dominant task. We evaluate the OCR performance of the first executable solution program in top-10 VLM predictions. \geovlm{}-OC, \geovlm{}-GC, \geovlm{}-GCD, and \geovlm{} achieve 1.84\%, 20.26\%, 13.95\%, and 46.58\% accuracy, respectively.
The result explains the accuracy improvement of \geovlm{} in the vision-dominant task against other variants.

\begin{table}[t!]
    \centering
    \resizebox{\linewidth}{!}{
    \begin{tabular}{l rr c rr}
        \toprule
        \multirow{2}{*}{Models} & \multicolumn{2}{c}{PGPS9K} && \multicolumn{2}{c}{GeoQA} \\
        \cmidrule{2-3} \cmidrule{5-6}
        & MR $\downarrow$ & mAP $\uparrow$ && MR $\downarrow$ & mAP $\uparrow$\\
        \midrule
        OpenCLIP & 50.50 & 27.87 && 111.70 & 1.29 \\
        GeoCLIP & 88.99 & 17.61 && 128.73 & 1.05 \\
        GeoCLIP-D & 58.83 & 13.35 && 107.25 & 2.86 \\
        GeoCLIP-DA & \textbf{12.88} & \textbf{41.13} && \textbf{35.60} & \textbf{33.25} \\
        \bottomrule
    \end{tabular}
    }
    \caption{Domain adaptation analysis. We report the mean rank (MR) and mean average precision (mAP) of the test diagrams. }
    \label{tab:domain_adaptation}
    \vskip -0.1in
\end{table}

\paragraph{Domain adaptation analysis.}
We examine how effectively GeoCLIP-DA generalizes to new domains with different diagram styles. For this experiment, we compare the embedding similarity between two diagrams representing the same structure in different styles. To create the paired dataset, we use a similar process described in \cref{sec:domain_adaptation}. Specifically, a total of 100 diagrams are sampled from the test sets of GeoQA and PGPS9K, and these samples are rendered in AlphaGeometry style through the diagram description.

For evaluation, we sample 100 diagrams from each of the target domain's training sets and compare the similarity against the original diagram via cosine similarity. We also compute the similarity between the style-transferred diagram and the original diagram. 
We report two metrics for test diagrams: the mean rank (MR) and the mean average precision (mAP) of the style-transferred diagram.

As reported in \cref{tab:domain_adaptation}, \geoclip{}-DA produces similar embeddings for structurally equivalent diagrams, regardless of their stylistic differences. \cref{fig:tsne} visualizes the diagram embeddings of OpenCLIP and GeoCLIP-DA. As one can observe, the OpenCLIP embeddings are largely separated by the domain of the diagrams, whereas those of \geoclip{}-DA appear to capture and align with underlying visual features more effectively.

\section{Conclusion}

In this work, we propose a domain-agnostic PGPS method, \geovlm{}, by implementing a synthetic data engine and proposing a contrastive learning framework with domain adaptation.
We demonstrate the effectiveness of \geovlm{} in visual feature perception at both VLM and vision encoder levels by evaluating on the MathVerse and through a newly proposed geometric feature recognition benchmark for vision encoders.
Eventually, the reasoning ability in plane geometry problems is enhanced with the improved perceptual capabilities.

\newpage
\section*{Limitations}

In this work, we present a domain-agnostic VLM for PGPS by refining the vision encoder. Although our VLM performs strongly in recognizing visual features, its coverage remains limited to geometric premises.
Building on the success of the synthetic data engine and contrastive learning, extending this combination to different kinds of visual features, e.g., sub-structures in molecular graphs~\citep{molvf}, statistics from charts~\citep{chartqa}, and solid geometry, promises further improvements in recognition of VLM. Due to the limitations in the experimental environment, we are unable to test LLMs with more than 30B parameters.

\section*{Acknowledgments}


This work was supported by the National Research Foundation of Korea(NRF) grant funded by the Korea government (MSIT) (No. RS-2024-00337955, RS-2023-00217286), Institute of Information \& communications Technology Planning \& Evaluation (IITP) grant funded by the Korea government (MSIT) (No. RS-2024-00457882, National AI Research Lab Project), and Institute of Information \& communications Technology Planning \& Evaluation (IITP) grant funded by the Korea government (MSIT) (No.RS-2019-II191906, Artificial Intelligence Graduate School Program (POSTECH))



\bibliography{custom}

\clearpage
\appendix
\renewcommand{\thefigure}{A\arabic{figure}}
\setcounter{figure}{0}
\renewcommand{\thetable}{A\arabic{table}}
\setcounter{table}{0}

\section*{Appendix}
\section{Details of Related Work}
\label{appendix:related_work}

\subsection{Comparison between MAVIS and ours}

Our contributions differ from MAVIS primarily in three aspects. 
Firstly, we introduce a systematic benchmark specifically designed for the quantitative analysis of vision encoders' capabilities in understanding geometric diagrams. 
This benchmark enables a fine-grained recognition evaluation across distinct geometric features, which is not addressed in MAVIS.

Secondly, we investigate the influence of caption style on vision encoder training by explicitly comparing GeoCLIP-style captions with MAVIS-style captions. 
MAVIS similarly employs extensive geometric diagram-caption pairs, namely MAVIS-Caption, where the captions include detailed geometric attributes such as object shape and connectivity. However, our empirical results comparing GeoCLIP and GeoCLIP (F ×), where GeoCLIP (F ×) uses captions incorporating all possible geometric premises, demonstrate that redundant geometric information within captions negatively impacts the vision encoder's recognition performance. Moreover, \cref{tab:linear_probing} shows that fine-tuning OpenCLIP with MAVIS-Caption with the setting as MAVIS yields significantly poorer performance on our visual geometric premise recognition benchmark compared to GeoCLIP and OpenCLIP.

Finally, our work explicitly addresses the issue of domain shift across different diagram styles by proposing a few-shot domain adaptation technique, a critical problem not considered in MAVIS.
\section{Synthetic Data Engine}
\begin{figure*}[t!]
    \centering
    \includegraphics[width=.9\linewidth]{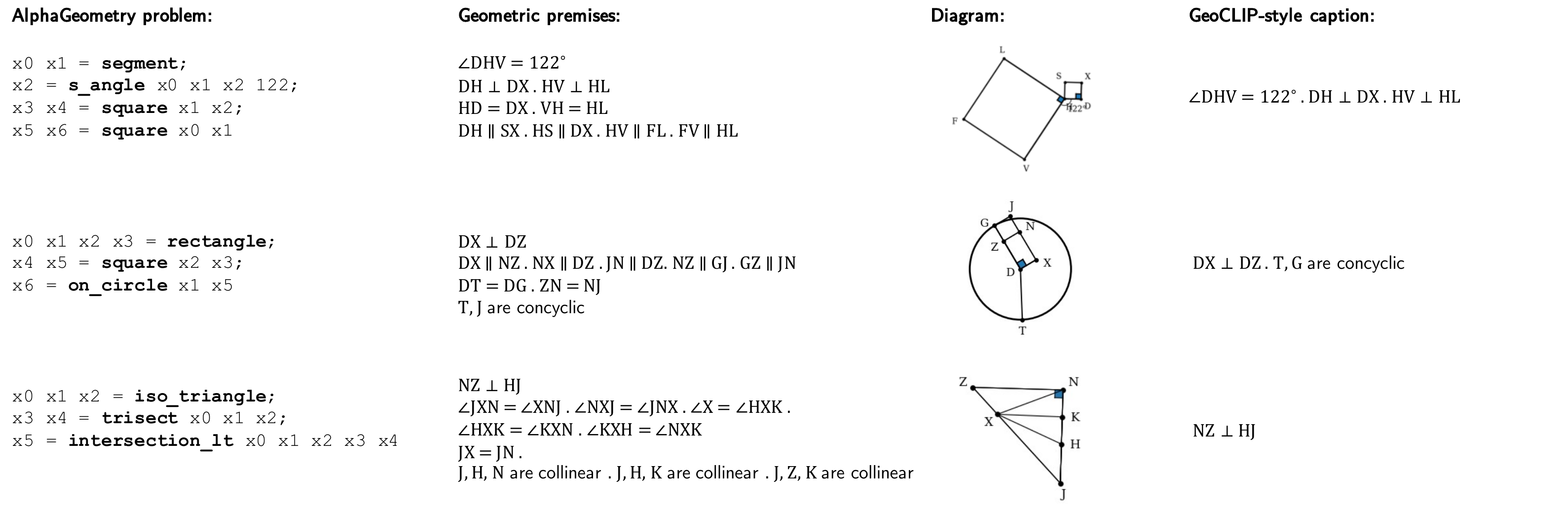}
    \caption{Example of randomly sampled AlphaGeometry problems. For each row, the first element describes the randomly sampled AlphaGeometry problem and the others are the geometric premises, diagram, and GeoCLIP-style caption that can be obtained from the AlphaGeometry problem. Note that the GeoCLIP-style caption can be obtained by filtering certain geometric properties, e.g., angle measure, perpendicularity, and concyclicity, from the geometric premises.}
    \label{fig:alphageometry}
\end{figure*}

\begin{table}[t!]
    \centering
    \begin{tabular}{l l}
    \toprule
    Visual premises & Non-visual premises \\
    \midrule
    \tabitem Perpendicularity & \tabitem Middle point \\
    \tabitem Collinearity & \tabitem Congruency in degree \\
    \tabitem Concyclicity & \tabitem Congruency in length \\
    \tabitem Parallelity & \tabitem Congruency in ratio \\
    \tabitem Angle measure & \tabitem Triangle similarity \\
    \tabitem Length measure & \tabitem Triangle congruency \\
    & \tabitem Circumcenter \\
    & \tabitem Foot \\
    \bottomrule
    \end{tabular}
    \caption{Geometric premises used in AlphaGeometry. \emph{Visual premises} denotes the geometric premises which can be directly perceived from the diagram. \emph{Non-visual premises} requires reasoning to be recognized.}
    \label{tab:alphageometry}
\end{table}

\begin{figure*}
    \centering
    \includegraphics[width=\linewidth]{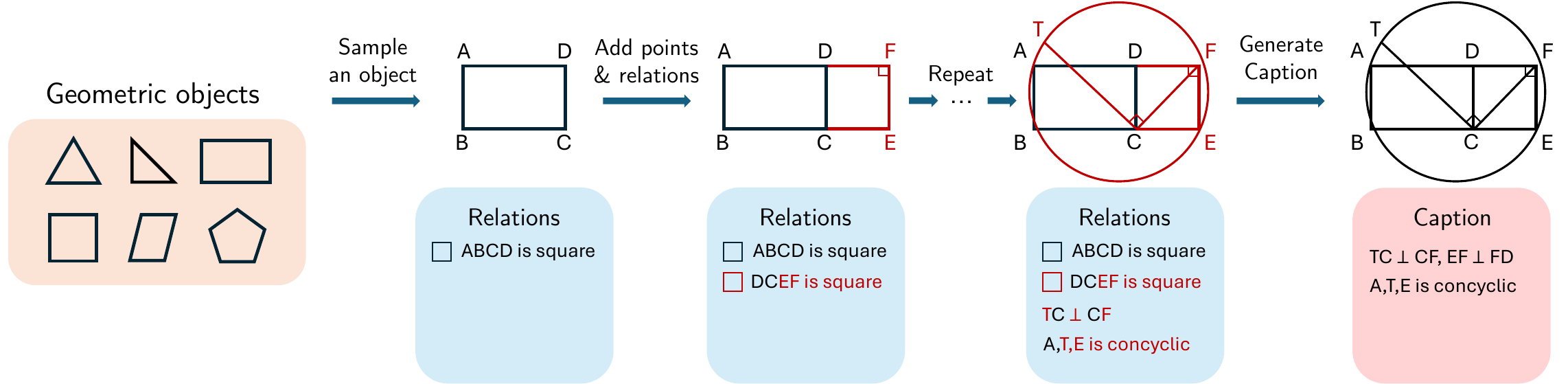}
    \caption{Illustration of the synthetic diagram-caption pairs generation. We first sample an object from a predefined object set. We then iteratively add points and relations to the existing primitives and relations. Finally, we generate a GeoCLIP-style caption based on the resulting primitives and relations.}
    \label{fig:data_synthesis}
\end{figure*}

In this section, we provide the details of our synthetic data engine. Based on AlphaGeometry~\citep{alphageometry}, we generate synthetic diagram and caption pairs by randomly sampling an AlphaGeometry program with \cref{alg:sampling}.
We visualize the AlphaGeometry program and diagram-caption pairs generation process in \cref{fig:data_synthesis}.

\begin{algorithm}[t!]
\caption{Sampling process of the synthetic data engine}
 \textbf{Input} Geometric relations $R$, geometric objects $O$, number of clauses $n_c$ \\
 \textbf{Output} AlphaGeometry program $c$
\begin{algorithmic}[1]
\State Initialize points and clauses with the sampled object: $P, C \sim O$ 
\For{$i \gets 1$ to $n_c$}
    \State Generate points: $P_{\text{new}}$
    \State Sample relation and points: $r, P_{\text{old}} \sim R, P$
    \State Construct clause: $C_{\text{new}} = r(P_\text{new}, P_\text{old})$
    \State Update points and clauses: $P, C \gets P \cup P_{\text{new}}, C \cup C_{\text{new}}$
\EndFor
\State Generate program with points and clauses: $c \gets \text{Clauses2Program}(P, C)$
\State \textbf{return} $c$
\end{algorithmic}
\label{alg:sampling}
\end{algorithm}

Examples for randomly sampled AlphaGeometry problems and their corresponding diagrams and lists of geometric premises are described in \cref{fig:alphageometry}.
The types of geometric premises that appear in our synthetic data engine are listed in \cref{tab:alphageometry}.

\section{Details of Benchmark}




\subsection{Role of the textual description}\label{sec:role_textual_description}

During the evaluation of the vision encoders, only the visual diagram serves as input to the model, and no textual information is provided during training or inference. Specifically, the evaluation is conducted using a linear probing approach, wherein the parameters of the vision encoder remain frozen, and only a linear classifier, initialized randomly, is trained atop the visual embeddings produced by the encoder.

Here, the textual questions from our benchmarks are implicitly represented in the classification labels assigned to each diagram. For instance, the textual question "How are lines AB and BC related?" corresponds directly to classification labels such as "Perpendicular," "Collinear," or "Neither." These labels are used as supervision signals to train the linear classifier. Thus, while the vision encoder receives no explicit textual input, the questions' semantics are reflected indirectly through the classification labels.

\subsection{Training details}\label{sec:hparams}

To evaluate the visual feature perception of the vision encoder, we utilize a linear probing approach, which involves freezing the vision
encoder parameters and training a simple linear classifier on top of its features.

We train the linear classifier on the training set of each task for 50 epochs with batch size 128 and learning rate $1\text{e-}4$.
We use Adam optimizer for optimization.

\subsection{Visualization of the vision encoders}

\begin{figure}[t!]
    \centering
    \begin{subfigure}[t]{.32\linewidth}
        \centering
        \includegraphics[width=\linewidth]{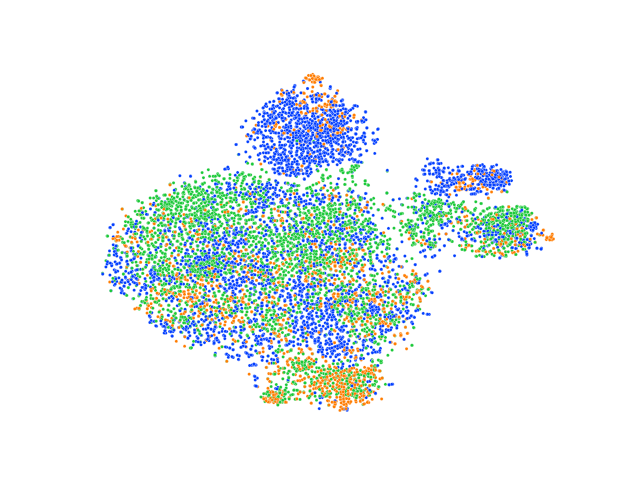}
        \caption{OpenCLIP}
    \end{subfigure}
    \begin{subfigure}[t]{.32\linewidth}
        \centering
        \includegraphics[width=\linewidth]{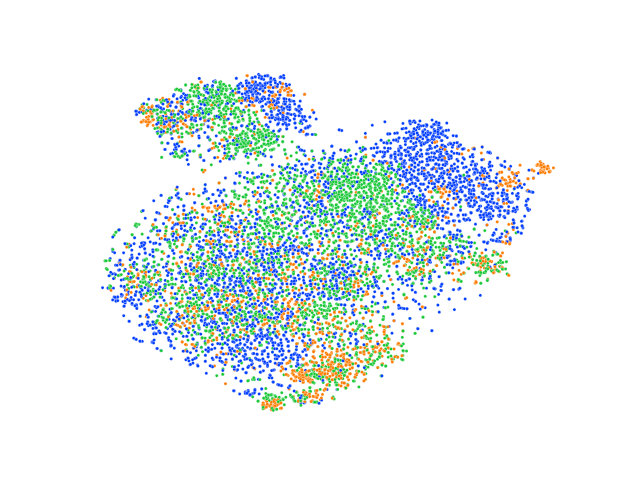}
        \caption{SigLIP}
    \end{subfigure}
    \begin{subfigure}[t]{.32\linewidth}
        \centering
        \includegraphics[width=\linewidth]{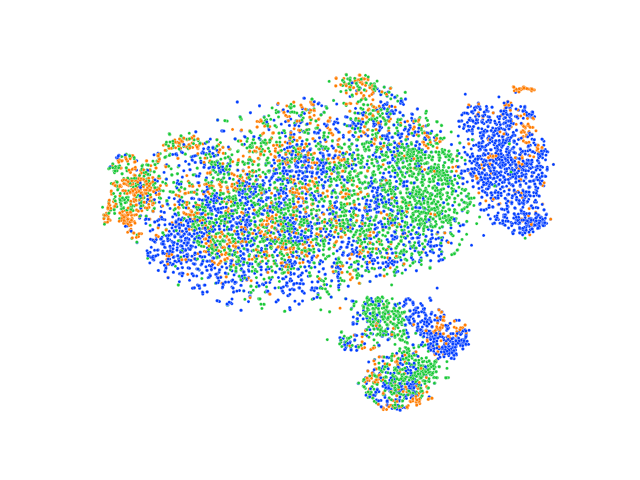}
        \caption{ConvNeXT}
    \end{subfigure}
    \begin{subfigure}[t]{.32\linewidth}
        \centering
        \includegraphics[width=\linewidth]{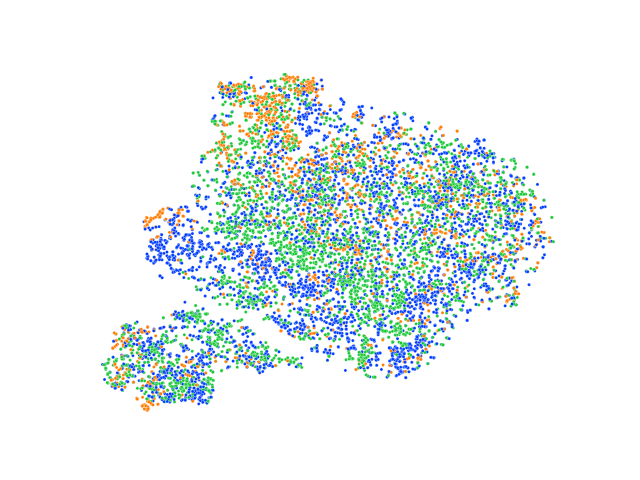}
        \caption{DinoV2}
    \end{subfigure}
    \begin{subfigure}[t]{.32\linewidth}
        \centering
        \includegraphics[width=\linewidth]{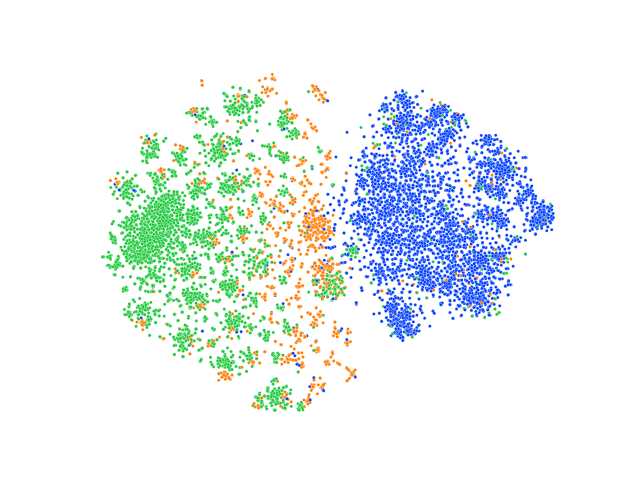}
        \caption{GeoCLIP}
    \end{subfigure}
    \caption{
    The embeddings of the vision encoders on the diagrams of TwoLines task. We visualize the embeddings of the vision encoders on the diagrams of TwoLines task. The blue, orange, and green dots are the diagrams where the two lines AB and BC are collinear, perpendicular, and otherwise, respectively.
    \label{fig:tsne_benchmark}
    }
\end{figure}

We visualize the embeddings of the vision encoders used in \cref{sec:benchmakr_results} at \cref{fig:tsne_benchmark}.

\section{GeoCLIP-DA}

\subsection{Details of the translation process}\label{sec:translation_details}

Note that the formal language is mentioned solely as an optional tool to accelerate the translation process when available, but it is not essential. The primary purpose of diagram translation in our method is to generate synthetic diagrams that visually resemble real-world diagrams. Since the objective is visual feature alignment rather than formal semantics, diagrams can effectively be translated manually by simply recreating the visual structure.

While manual translation might appear time-consuming, the effort is minimal and feasible in practice. Specifically, we manually translated only around 50 diagrams per domain, which required less than 3 hours in total. This modest effort substantially improved our model's cross-domain generalization performance. Thus, manual translation without formal annotations is not only practical but also highly beneficial for domain adaptation. The process of the translation is illustrated in \cref{fig:translation_process}

\begin{figure}[t!]
    \centering
    \includegraphics[width=\linewidth]{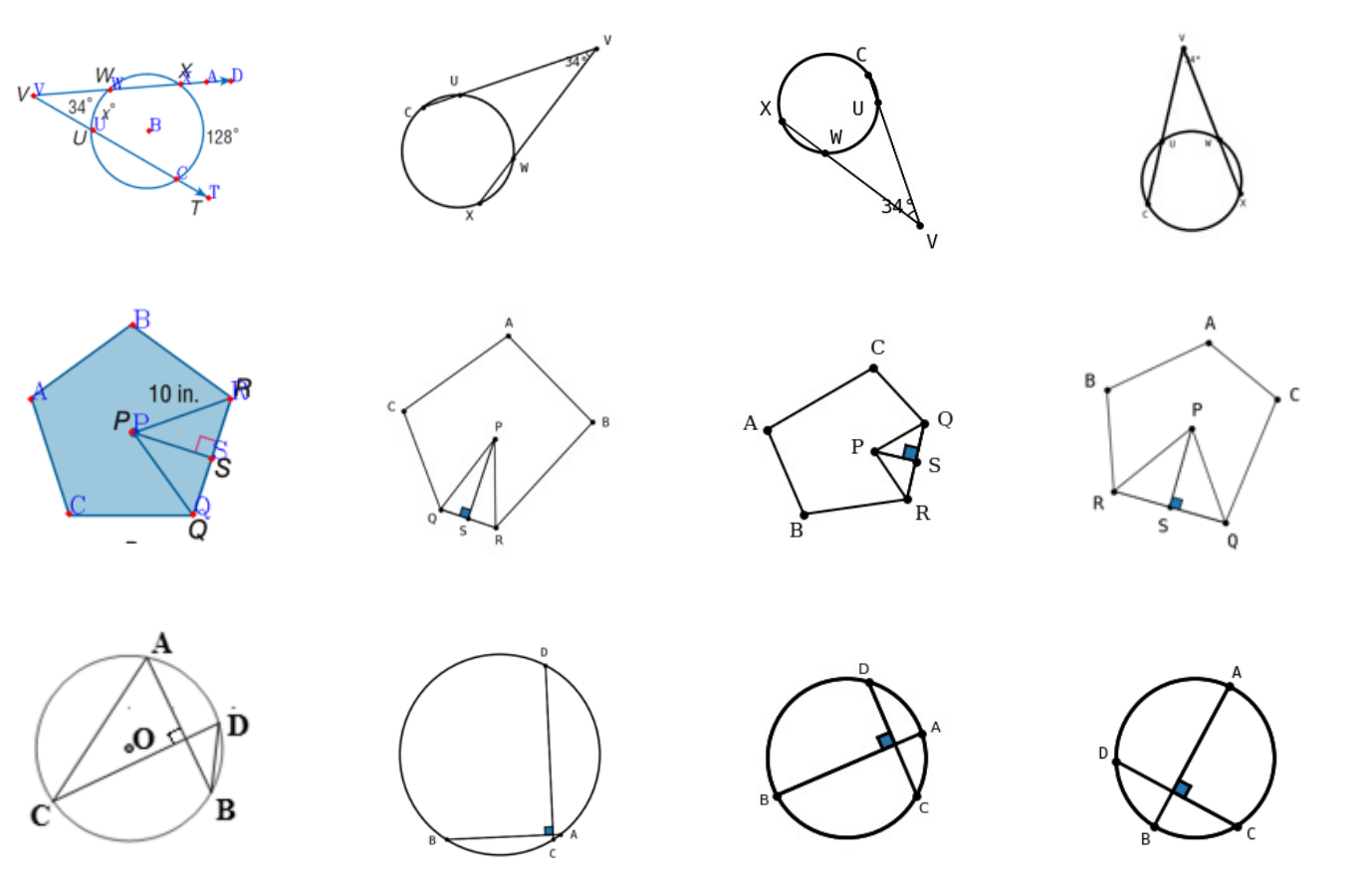}
    \caption{Examples of diagram pairs curated for domain adaptation. For each row, the first diagram is from the target domain, and the remaining diagrams are from the source domain. To generate source domain diagrams, we translate the target diagram by our diagram generator with the textual description of the target image.}
    \label{fig:domain_adaptation_samples}
\end{figure}

\subsection{Domain adaptation data}

We adopt GeoCLIP to the two PGPS benchmarks: GeoQA~\citep{geoqa} and PGPS9K~\citep{pgps}.
For PGPS9K, we use the Geometry3K split.
\cref{fig:domain_adaptation_samples} shows the pairs used to adapt the domain of GeoCLIP.

\subsection{Training details}

We start from OpenCLIP~\citep{clip}, a pre-trained model where the architecture is ViT-L/14 with image resolution $336\times 336$. To train OpenCLIP, we use total of 200,000 diagram-caption pairs generated with our synthetic data engine.
For the domain adaptation to GeoQA and Geometry3K datasets, we randomly sample 50 diagrams and translate the diagram and caption styles following the procedure described in \cref{sec:domain_adaptation}. Finally, \geoclip{} is fine-tuned via \cref{eqn:da}.
We name the GeoQA and Geometry3K adopted \geoclip{} as \geoclip{}-DA.

We set the batch size for the source domain diagram-caption pairs to 256. 
For the domain adaptation parts, i.e., applying CLIP on the diagram-caption pairs and the diagram pairs of target domains, we vary the batch size to 32.
We set weight decay to 0.2.
We optimize for 50 epochs using Adam optimizer~\citep{adam} and a cosine annealing scheduler with 2,000 warmup steps, and the maximum learning rate is set to be $1\text{e-}4$.
We train the model with eight RTX3090 GPUs for approximately 24 hours.

\subsection{Visualization of GeoCLIP-DA embeddings}

\begin{figure}[h!]
    \centering
    \includegraphics[width=.92\linewidth]{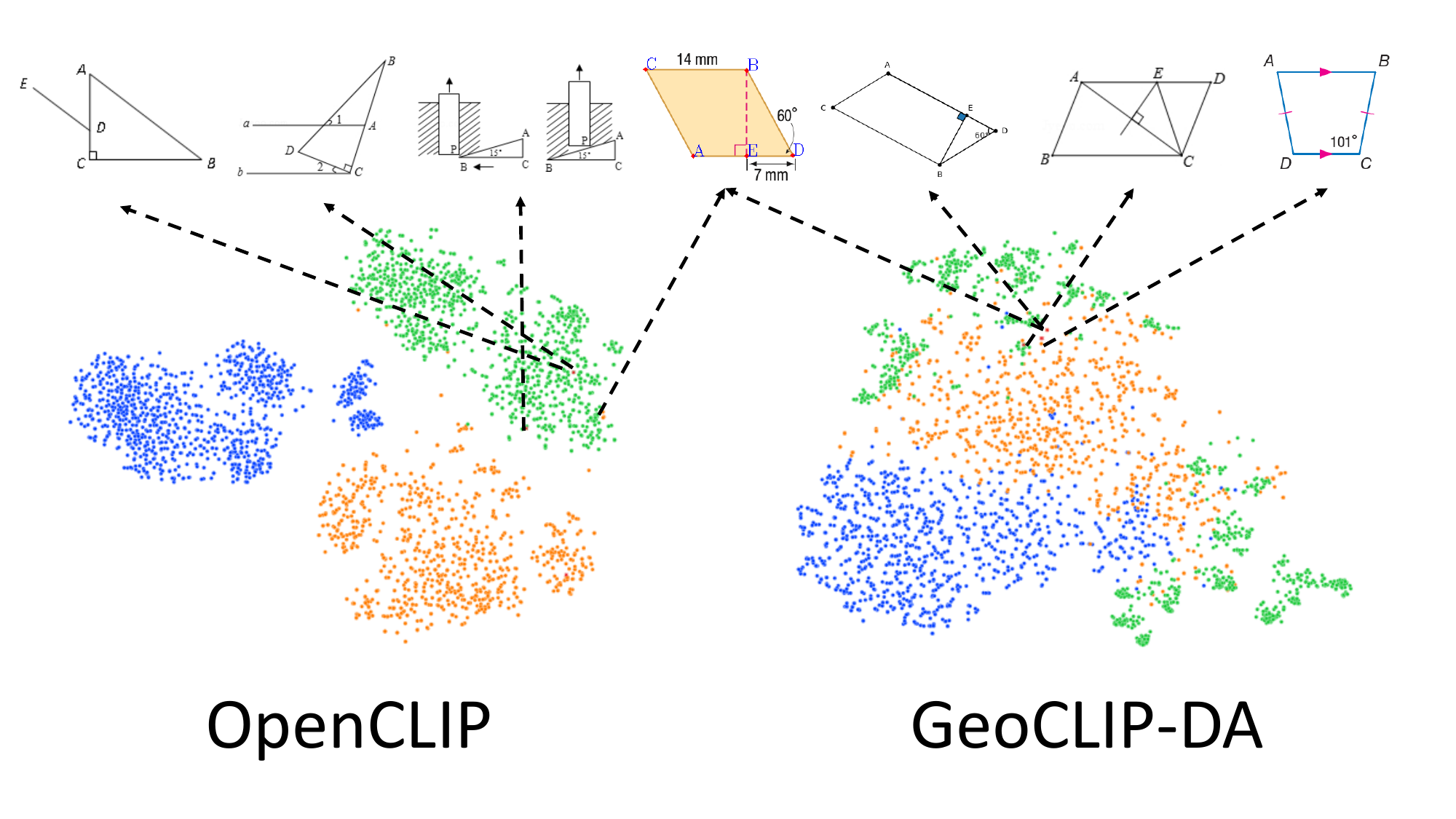}
    \caption{Visualization of OpenCLIP and GeoCLIP-DA embeddings. The orange, green, and blue dots represent PGPS9K, GeoQA, and synthetic diagrams, respectively. In the top row, the three diagrams on the left and right are those with the highest cosine similarities to the center under OpenCLIP and GeoCLIP-DA, respectively.}
    \label{fig:tsne}
    \vskip -0.19in
\end{figure}

We compare the embeddings between GeoCLIP-DA and OpenCLIP in \cref{fig:tsne}.

\section{\geovlm{}}
\label{sec:vlm_details}

\begin{figure}[t!]
    \centering
    \includegraphics[width=\linewidth]{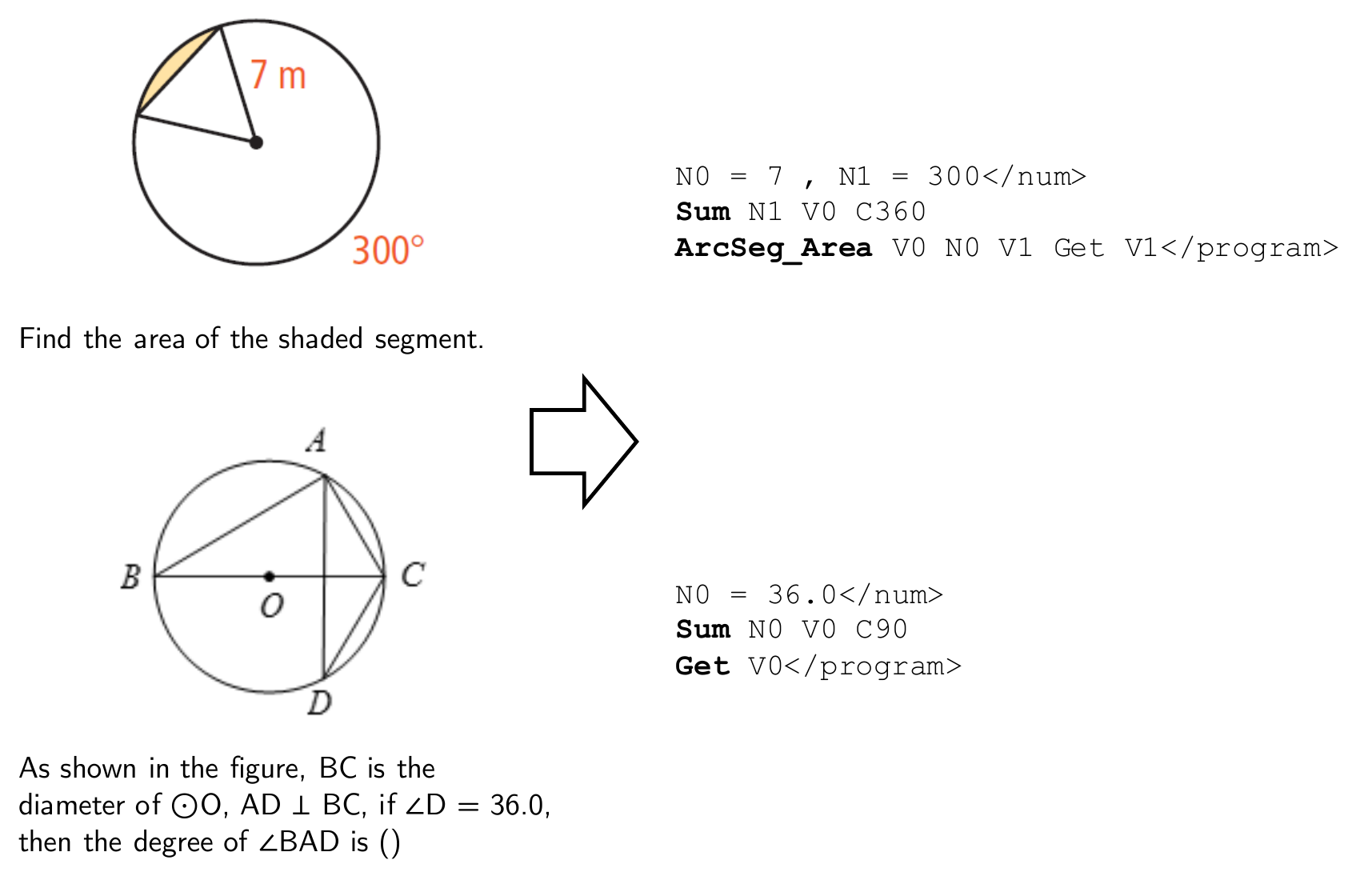}
    \caption{Examples of the training data for GeoDANO. While previous PGPS models require only predicting the solution steps and assuming the numerical values are explicitly given, GeoDANO is trained to predict both the solution steps and the numerical values in the diagram and text.}
    \label{fig:training_data}
\end{figure}

\subsection{Training details}
\paragraph{Architectural details.}
We begin by summarizing the architecture of our VLM, a combination of a vision encoder and a language model. For the vision encoder, we use \geoclip{}-DA, with a two-layer MLP of GeLU activation as the projection layers following LLaVA-OneVision~\citep{llava-next}. For the language model, we employ LLama-3-8B-Instruct~\citep{llama}.
For a given diagram and question pair in PGPS, we feed the vision encoder with the given diagram, and then the output of the encoder is used as an input token of LLM through the projection layer. The question text is then fed into the LLM, followed by the diagram embedding.

\paragraph{Training approach.}
With the modified training data, we apply supervised fine-tuning on the VLM, i.e., the gradient only flows through the prediction of numerical values and solution steps, not the diagram and text.
During the training of GeoDANO, the parameters of the vision encoder, i.e., GeoCLIP-DA, are frozen and remain unchanged. The projection layer, which maps visual embeddings to language model inputs, is randomly initialized and trained from scratch simultaneously with the language model.

\paragraph{Hyper-parameters.}
We train the VLM with AdamW optimizer~\citep{adamw} and cosine annealing scheduler with warmup ratio 0.03 and maximum learning rate $1\text{e-}5$.
We use LoRA~\citep{lora} with rank 128.
We set the batch size to 16 and train with 5 epochs.
We train the VLM with four A100-80GB GPUs for approximately 24 hours.






\end{document}